\documentclass[journal]{IEEEtran}

\usepackage{cite}
\usepackage[T1]{fontenc}
\usepackage{amsmath,amssymb,amsfonts}
\usepackage{algorithmic}
\usepackage{graphicx}
\usepackage{textcomp}
\usepackage{xcolor}
\usepackage{wrapfig,epsfig,epstopdf}
\usepackage{subfigure}
\usepackage{float}
\usepackage{array}
\usepackage{makecell}
\usepackage[ruled]{algorithm}
\usepackage[english]{babel}
\usepackage{comment}
\usepackage{tabularx}
\usepackage{rotating,multirow}
\usepackage{booktabs}
\usepackage[subnum]{cases}

\newtheorem{theorem}{Theorem}
\usepackage{array}

\begin{document}

\title{Active Multi-Label Crowd Consensus}

\author{Jinzheng~Tu, Guoxian~Yu, Carlotta~Domeniconi, Jun~Wang, Xiangliang~Zhang
\thanks{J.~Tu, G.~Yu and J.~Wang are with the College of Computer and Information Sciences, Southwest University, China (e-mail: \{tujinzheng, gxyu, kingjun\}@swu.edu.cn)}
\thanks{C.~Domeniconi is with  the Department of Computer Science, George Mason University, USA (e-mail: carlotta@cs.gmu.edu)}
%\thanks{M.~Guo is with the School of Electrical and Information Engineering, Beijing University of Civil Engineering and Architecture, China (e-mail: guomaozu@bucea.edu.cn)}
\thanks{X.~Zhang is with the King Abdullah University of Science and Technology, Thuwal, SA (e-mail: xiangliang.zhang@kaust.edu.sa)}
\thanks{Guoxian Yu are the corresponding author, gxyu@swu.edu.cn.}
%\thanks{Manuscript received \today; Revised XXXX}
}

\maketitle

\begin{abstract}
Crowdsourcing is an economic and efficient strategy aimed at collecting annotations of data through an online platform. Crowd workers with different expertise are paid for their service, and the task requester usually has a limited budget. How to collect reliable annotations for multi-label data and how to compute the consensus within  budget is an interesting and challenging, but rarely studied, problem.

In this paper, we propose a novel approach to accomplish Active  Multi-label Crowd Consensus (AMCC). AMCC accounts for the \emph{commonality} and \emph{individuality} of workers, and assumes that workers can be organized into different groups. Each group includes a set of workers who share a similar annotation behavior and label correlations. To achieve an effective multi-label consensus, AMCC models workers' annotations via a linear combination of commonality and individuality, and reduces the impact of unreliable workers by assigning smaller weights to the group. To collect reliable annotations with reduced cost, AMCC introduces an active crowdsourcing learning strategy that selects \emph{sample-label-worker} triplets. In a triplet, the selected sample and label are the most informative for the consensus model, and the selected worker can reliably annotate the sample with low cost. Our experimental results on multi-label datasets demonstrate the advantages of AMCC over state-of-the-art solutions on computing crowd consensus and on reducing the budget by choosing cost-effective triplets.
\end{abstract}


\IEEEpeerreviewmaketitle

\section{Introduction}
%Crowdsourcing is an idea of outsourcing a task to a large group of networked people in the form of an open call to reduce the production cost \cite{Howe2006The}.
%Tasks that are rather trivial for humans, but unsolvable for machines (e.t., sentiment classification \cite{snow2008cheap}, image tagging \cite{Welinder2010The}, medical diagnosis \cite{Raykar2010Learning}) can be efficiently addressed by crowdsourcing. Many  crowdsourcing platforms, such as Amazon Mechanical Turk\footnote{http://www.amt.com}, CrowdFlower\footnote{http://www.crowdflower.com} and Baidu Test\footnote{http://test.baidu.com/crowdtest/},  are widely-used for various crowdsourcing tasks.

Crowdsourcing is the practice of collecting information from a large group of people in the form of an open call to reduce  production costs \cite{li2016crowdsourced}.
Tasks that are rather trivial for humans, but difficult for machines (e.g., sentiment classification \cite{snow2008cheap}, image tagging \cite{Welinder2010The}, and medical diagnosis \cite{Raykar2010Learning}) can be efficiently addressed using crowdsourcing. Many  crowdsourcing platforms, such as Amazon Mechanical Turk\footnote{http://www.amt.com}, CrowdFlower\footnote{https://www.figure-eight.com}, and Baidu Test\footnote{http://test.baidu.com/crowdtest/},  are widely-used for various crowdsourcing tasks. Annotations from the crowd always contain many noisy labels, which have been attracting increasing attention in various domains \cite{duan2016learning,han2018progressive,varon2015noise,zhu2015constrained,miao2015rboost}.

Computing the crowd consensus annotation from repeated annotations provided by multiple workers on the same take is the key issue in crowdsourcing \cite{kazai2012face}. Many consensus algorithms have been suggested, each pursuing a different criterion (i.e.,  reliability \cite{Whitehill2009Whose}, intention \cite{venanzi2016time}, difficulty of samples \cite{kurve2015multicategory}, and bias of workers \cite{kamar2015identifying,Welinder2010The}). Computing crowd consensus in a multi-label problem is even more challenging due to the high order of possible available label combinations.

In this paper, we study an interesting and practical topic, {\bf active multi-label crowd consensus learning}, which aims at achieving reliable consensus labels with minimized budgets. Although active learning has been introduced to reduce the annotation cost by  selecting  the most valuable samples to be  queried \cite{settles09}, its potential and feasibility in multi-label crowd consensus learning has not been well studied, mainly because of the following \emph{challenges}. i) Traditional active learning generally employs an oracle or an expert (which in practice may not be available, or very expensive), and assumes that the provided annotations are correct. In contrast, the selected samples in active crowdsourcing learning are annotated by different \emph{non-reliable} workers, who may give incorrect annotations. ii) Traditional active learning focuses on samples, labels, costs, and sample-label pairs separately, or considers at most two at one time \cite{yan2012active,yan2011active,Huang2013Active,li2018multi,yan2018cost}, whereas active multi-label crowd consensus should jointly account for the workers (specialty and commonality), costs, samples, and labels. The latter aims at selecting useful but cost-saving  \emph{sample-label-worker} triplets for the query. (iii) Existing active crowdsourcing learning approaches \cite{Zhao2011Incremental,Dekel2016Selective,yan2012active,Donmez2009Efficiently,yan2011active} cannot be directly adapted for multi-label crowd consensus problems, and they either ignore label correlations, specialty of workers, or their costs. Some approaches still assume an ideal worker is selected for annotation \cite{Fang2014Active}. In practice, such worker does not exist, or is very expensive.

To address these intrinsic challenges, we introduce an active multi-label crowd consensus approach (called AMCC) for efficient and cost-saving crowdsourcing. AMCC assumes that the annotation matrix of a worker is a linear self-representation of two matrices: one describes the expertise (or individuality) of each worker, and the other encodes the commonality of a group of workers. Workers in the same group tend to exhibit similar behavioral traits when annotating samples \cite{simpson2013dynamic}, so they have  similar reliability and bias, and share the same label correlations. For example, one group may only include reliable workers, while another group mainly includes spammers or workers with low reliability, whose annotations are less credible than the former. AMCC computes the crowd consensus using the commonality and individuality of workers, and assigning low weights to low-quality groups. In addition, considering the limited budget, different specialties and costs of individual workers, we further introduce an active crowdsourcing learning strategy to select the informative but cost-saving sample-label-worker triplets for the query. The main contributions of this paper are summarized as follows:
\begin{enumerate}
  \item We introduce a novel Active Mmulti-label Crowd Consensus approach (AMCC) to automatically bridge active learning with multi-label data crowdsourcing. AMCC jointly makes use of the intrinsic expertise of different groups of workers, inter-relations between workers, and label correlations of each group. It can assign different weights to these groups to further reduce the impact of low quality workers and to compute reliable consensus labels.
  \item The partition of workers into groups not only reduces the number of weight parameters and the impact of sparse annotations, but also contributes to explore intrinsic label correlations of each group.
  \item We introduce a novel active crowdsourcing learning strategy to select sample-label-worker triplets {based on workers' expertise and costs}, in such a way that the selected samples and labels can best improve the consensus model, and the selected worker can  annotate the sample in a reliable manner and at a low cost. To the best of our knowledge, \emph{none} of the existing active crowdsourcing solutions \cite{Zhao2011Incremental,Dekel2016Selective,yan2012active,Donmez2009Efficiently,yan2011active,Fang2014Active} can jointly account for the impact of samples, labels, and workers in crowdsourcing.
  \item Extensive results validate the advantages of our proposed AMCC approach over  state-of-the-art solutions \cite{Duan2015Crowdsourced,zhang2018multi,Yoshimura2017Quality,Tu2018multi,li2018multi} in effectively computing  the multi-label crowd consensus and saving costs.
\end{enumerate}

The remainder of this paper is organized as follows. Section \ref{sec:relwork} briefly reviews multi-label crowd consensus learning and active crowdsourcing learning. In Section \ref{AMCC}, we discuss the computation of multi-label crowd consensus and the selection of sample-label-worker triplets for active learning. Section \ref{exp} provides the experimental setup and results. Section \ref{concl} gives conclusions and future work.

\section{Related Work}
\label{sec:relwork}
In this section, we briefly review two branches of crowdsourcing: \emph{quality control} and \emph{active learning}, which have close connections with our work.

\paragraph{Quality Control} In a crowdsourcing platform, people with different backgrounds annotate data in exchange of a typically modest reward \cite{Kazai2011Crowdsourcing}. A data object can be annotated with one (or more) label(s)  by several workers, based on their knowledge \cite{Sheng2008Get}. The sample-label information collected via crowdsourcing is generally erroneous, due to the fact that online workers may lack expertise and proper incentives \cite{Kazai2011Crowdsourcing,vuurens2011much}.  This heterogeneous nature leads to the diverse submission quality of the completed tasks, pressing an urgent need for quality control \cite{daniel2018quality,zhang2019crowdsourced,zhang2017improving,allahbakhsh2013quality,hu2019quality,nowak2010reliable,hung2018computing,Duan2015Crowdsourced,Tu2018multi,zhang2018multi}.

\textbf{Computing crowd consensus} in a reliable manner, such as eliminating low quality workers and spammers\cite{Tu2018multi}, is a key issue in crowdsourcing \cite{kazai2012face}.
%Many consensus algorithms have been suggested, each pursuing a different criterion (i.e.,  reliability \cite{Whitehill2009Whose}, intention \cite{venanzi2016time}, difficulty of samples \cite{kurve2015multicategory}, and bias of workers \cite{kamar2015identifying,Welinder2010The}).
Existing consensus algorithms \cite{Whitehill2009Whose,venanzi2016time,kurve2015multicategory,kamar2015identifying,Welinder2010The} can produce integrated labels with improved quality. However, they all focus on binary scenarios. As a result, they may perform poorly when dealing with the more general multi-label data setting, where each object may have a set of non-exclusive labels, and labels may exhibit  semantic correlations. Several multi-label crowd consensus algorithms have been recently proposed \cite{nowak2010reliable,hung2018computing,Duan2015Crowdsourced,Tu2018multi,zhang2018multi}. Nowak \textit{et al.} \cite{nowak2010reliable} studied the inter-annotator agreement for multi-label image annotation and focused on the annotation quality differences between expert and non-expert workers. Duan \textit{et al.} \cite{Duan2015Crowdsourced} introduced a probabilistic cascaded method (C-DS) to compute multi-label crowd consensus. These solutions ignore the correlation between labels, whose appropriate usage can improve the consensus labels and also reduce the budget \cite{Yoshimura2017Quality,zhang2018multi,Tu2018multi,chen2018cost}.  Yoshimura \textit{et al.} \cite{Yoshimura2017Quality} proposed RA$k$EL-GLAD to balance  estimation accuracy and computational complexity for computing multi-label crowd consensus. Hung \textit{et al.} \cite{hung2018computing} proposed a Bayesian non-parametric consensus approach, which extends the clustering based Bayesian combination of classifiers for multi-label data by additionally incorporating co-occurrence dependence between labels.  However, this Bayesian method asks for sufficient annotations of the training data. Furthermore, these multi-label consensus solutions \cite{Duan2015Crowdsourced,Yoshimura2017Quality} neglect to model the similarity between workers, which can improve the consensus. To remedy this issue, Zhang \textit{et al.} \cite{zhang2018multi} proposed a probabilistic multi-class multi-label dependency method (MCMLD) to model the reliability of workers using a set of confusion matrices. It captures label correlations using a mixture of multiple independent multinoulli distributions, and then computes the crowd consensus of each sample. However, these methods cannot identify the widely-witnessed spammers, who randomly (or identically) annotate the data to earn easy money. Tu \textit{et al.} \cite{Tu2018multi} introduced a joint matrix factorization based solution (ML-JMF) to optimize the weights of individual workers, and thus to identify spammers by crediting them zero weights. However, ML-JMF suffers from the common phenomenon of sparse annotations, which occurs when workers  annotate objects only with few out of the several  relevant labels.
%In addition, it suffers from a high number of weight parameters, due to  a large number of involved workers.

{Since workers have diverse qualities on different tasks, a better task assignment strategy may also contribute to a better consensus. Many researches on improving task design have been explored from different perspectives, such as lower complexity \cite{rogstadius2011assessment}, worker's expertise \cite{Zhang2017Expertise,tao2018domain}, checking workers' answers \cite{bragg2013crowdsourcing}, and so on. To name a few, Zhang \textit{et al.} \cite{Zhang2017Expertise} considered an expertise-aware task allocation problem in mobile crowdsourcing, where the worker's expertise is obtained based on semantic analysis. Rokicki \textit{et al.}\cite{rokicki2015groupsourcing} studied dynamically recruiting teams of workers for specific tasks.}

\paragraph{Active crowdsourcing}
In real-world applications, {the crowd is not free, if there are large numbers of tasks, crowdsourcing can still be expensive and time-consuming.} As a result, it is wise to collect reliable annotations for fewer but valuable samples to train an accurate consensus model. To this end, active learning can be incorporated. Active learning aims at reducing the annotation cost by  selecting  the most valuable samples to be  queried \cite{settles09}. In  canonical active learning, the annotations of the selected samples are obtained from an expert (or oracle), who is assumed to possess the ground truth \cite{Li2013Active}. In contrast, active learning in  crowdsourcing is much \emph{more challenging};  experts may not be available, and multiple non-expert workers can annotate the samples. As such, the real labels of samples  can only be approximated from the consensus of the workers.

Several approaches have been suggested for active crowdsourcing learning \cite{Zhao2011Incremental,Dekel2016Selective,yan2012active,Donmez2009Efficiently,yan2011active,Fang2014Active}. Zhao \textit{et. al} \cite{Zhao2011Incremental} combined the uncertainty and inconsistency measures to actively select the most important samples for re-annotation, but they could not identify appropriate annotators for the selected samples. Zheng \textit{et. al} \cite{zheng2010active}  collected the annotations of a subset of annotators, who are globally chosen for all samples. As a result, an unnecessary high cost may be introduced, since the expertise and cost of different annotators are ignored. Other approaches try to select annotators with matching expertise for individual samples to reduce the cost \cite{Dekel2016Selective,yan2012active,Donmez2009Efficiently,yan2011active,Fang2014Active}.
However, these approaches do not differentiate the costs of individual workers. As such, they may get an accurate, yet expensive solution.

Our proposed AMCC can jointly model the worker's specialty and commonality, costs, samples, labels in a unified model, and acquire reliable annotations from crowd workers with much reduced costs. The experimental results show that AMCC not only can obtain better consensus labels, but also lower costs than these related and competitive solutions \cite{Duan2015Crowdsourced,zhang2018multi,Yoshimura2017Quality,Tu2018multi,li2018multi}.

\section{Active Multi-Label Crowd Consensus}
\label{AMCC}
\subsection{Problem Formulation}
We consider active learning in the multi-label crowdsourcing setting, where a set of labeled samples  $\mathcal{D}^L=\{(\mathbf{x}_i, \mathbf{y}_i)\}_{i=1}^{N_l}$ and a large pool of unlabeled samples $\mathcal{D}^U=\{\mathbf{x}_j\}_{j=1+N_l}^{N_l+N_u}$ are  available. Typically, $N_l\ll N_u$.
Each sample $\mathbf{x}_i$ is assumed to have a collection of labels taken from the set $\mathcal{L}=\{1, 2, \ldots, L\}$, which includes $L$ distinct semantic labels.
The labels of samples in $\mathcal{D}^L$ are partially known, while the labels of samples in  $\mathcal{D}^U$ are all unknown. A collection of  $W$ workers, denoted as $\mathcal{W}=\{1,\ldots,W\}$, assigns labels from $\mathcal{L}$ to samples. Formally, each worker $w$ provides a sample-label association matrix $\mathbf{A}_w \in \mathbb{R}^{N_l \times L}$ for $N_l$ samples and $L$ labels. $\mathbf{A}_w(i,j)=a^w_{il} \in \left\{- 1,0,1 \right\}$. $a^{w}_{il}=1 (-1)$  states that the $w$-th worker annotated the $i$-th sample with (or without) the $l$-th ($1 \leq l \leq L$) label; $a^w_{il}=0$ means that the worker did not specify whether  the  $\mathbf{x}_i$ has the $l$-th label.

\begin{figure}\vspace{-0.3cm}
  \centering
  \includegraphics[width=8cm,height=4.5cm]{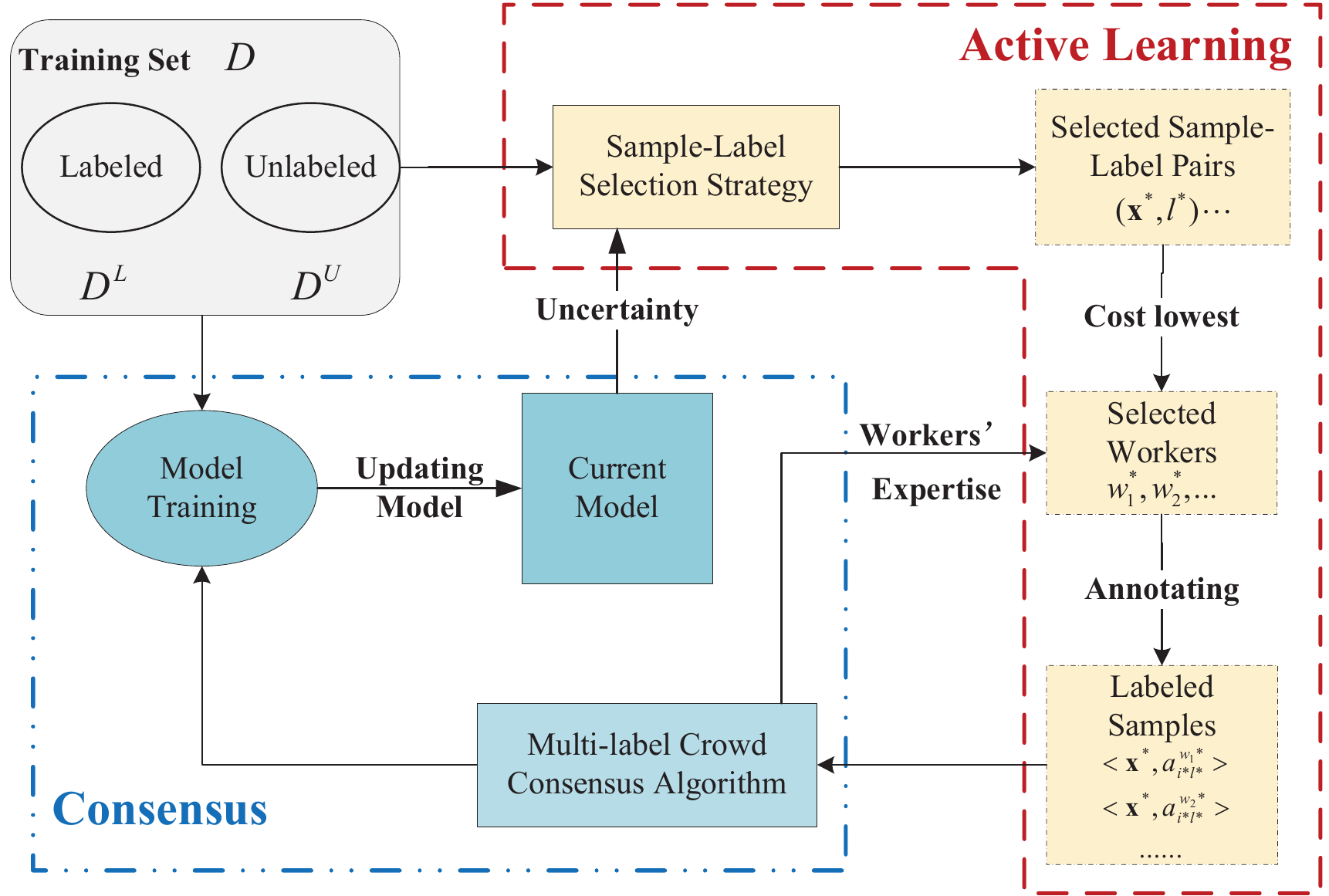}\\
  \vspace{-0.2cm}
  \caption{The framework of active multi-label crowd consensus, which consists of four steps: sample-label selection, worker selection, multi-label crowd consensus computation, and model updating.}\label{frame}\vspace{-0.5cm}
\end{figure}

The workflow of AMCC is illustrated in Fig. \ref{frame}. Our  sample-label-worker triplet selection strategy is first applied to select a group of sample-label pairs from $\mathcal{D}^L \bigcup \mathcal{D}^U$, that have a high degree of uncertainty, and a reduction of the latter is likely to improve the consensus.  Consequently, workers with the lowest cost but still capable of providing credible annotations for the selected samples and labels are chosen for the annotation. Then, our multi-label crowd consensus algorithm is triggered to update the consensus labels. The sample-label-worker selection and the multi-label consensus processes are iteratively executed until the query budget is exhausted, or the consensus model cannot be  further improved.

\subsection{Multi-label Crowd Consensus}
%In addition, there exists different types of workers. (1)reliable workers have high accuracy; (2)normal workers have relatively high accuracy, but not as accurate as reliable workers; (3) sloppy workers have low accuracy. their annotations are not so reliable in general; (4) random workers have low accuracy, they often carelessly annotate samples; (5) uniform spammers also have low accuracy, they uniformly (or randomly) annotate a label to all samples \cite{Tu2018multi}.
In crowdsourcing tasks, engaged workers can be different in age, gender, interest, and so on. It is also recognized that workers may share aspects of the annotation behavior, and they can be partitioned into groups accordingly \cite{kurve2015multicategory}. {Since the workers hold different expertise and share similar annotation behaviors,} to account for the individuality and commonality  among these workers, we assume that each annotation matrix is expressed as a linear combination of two matrices as follows:
\begin{equation}\label{eq1}
  \mathbf{A}_{w}= \mathbf{A}_{w}(\mathbf{D}_{w}+\mathbf{C}_{m})
\end{equation}
$\mathbf{D}_{w}\in \mathbb{R}^{L\times L}$ encodes the individuality of worker $w$ by measuring  the worker's capability of selecting the correct label over all label pairs.
{$\mathbf{D}_w(g,l)$ represents the probability that worker $w$ incorrectly selects label $l$ instead of label $g$ ($l\neq g$).}  $\mathbf{C}_m \in \mathbb{R}^{L\times L}$ encodes the commonality of the $m$-th group of workers. {Each element $\mathbf{C}_m(g,l)$ represents the probability that all the workers in the $m$-th group give label $l$ when the truth is $g$. In this way, label correlations of multi-label data are explored and considered.}

Since annotations were collected from multiple workers, we generalize the above equation as:
\begin{equation}\label{eq2}\small
\begin{split}
\min\limits_{\mathbf{D}_w, \mathbf{C}_m}\sum_{m=1}^M\sum_{w=1}^W\boldsymbol{\lambda}_m^r\|\mathbf{A}_w-\mathbf{A}_w
  (\mathbf{D}_w+\mathbf{C}_m)\|^2_F \\
  +\alpha\Omega(\mathbf{D}_1, \cdots, \mathbf{D}_W)+\beta\Psi(\mathbf{C}_1, \cdots, \mathbf{C}_M)\\
  s.t. \ \
 \sum_{l=1}^L\mathbf{D}_w(g,l)=1,\ \sum_{m=1}^M \boldsymbol{\lambda}_m=1
\end{split}
\end{equation}
where $\left\|  \cdot  \right\|_F^2$ is the Frobenius norm, $M$ is the specified number of groups. $\boldsymbol{\lambda}_m>0$ is used to automatically weigh different groups, and to reduce the impact of low quality groups and workers therein. The scalar parameter $r>1$ is added to avoid considering only one group. The second term, $\Omega(\mathbf{D}_1, \ldots, \mathbf{D}_W)$, enables the assignment of  inter-dependent and similar workers to the same group. The last term, $\Psi(\mathbf{C}_1, \cdots, \mathbf{C}_M)$ pushes each group to have its own commonality signature and label correlations, which also reflect the bias of workers in the group  in annotating samples.  In this way, workers are partitioned into different coherent groups, and the latent label correlations and connections between workers are implicitly encoded. $\alpha$ and $\beta$ are  scalar parameters that balance the importance of the two terms. We observe that grouping workers not only reduces the number of weight parameters (from $W$ to $M$), and the impact of sparse annotations by merging annotations of workers of the same group, but also contributes to the exploration of intrinsic label correlations of each group, and to the subsequent active crowdsourcing learning. Our experimental results corroborate the advantages of grouping workers.

Workers with similar annotation behaviors should be assigned to the same group. Our objective is to find groups that maximize the inter-dependence between the member workers. Classic measures of correlation include Spearmans rho and Kendall tau \cite{yue2002power}, but they can only detect linear dependency. We employ the Hilbert-Schmidt Independence Criterion (HSIC) \cite{gretton2005measuring} to quantify the dependence between $\mathbf{D}_w$ and $\mathbf{D}_v$. We use HSIC because it can measure both linear and nonlinear dependences. In addition, it estimates dependence between variables without explicitly estimating their joint distribution. As a result, it's  computational efficient. Furthermore, the empirical HSIC is equal to the trace of the data matrix product, which makes our problem solvable.

Suppose $\phi(\mathbf{x})$ maps $\mathbf{x}$ onto a kernel space $\mathcal{F}$ such that the inner product between vectors in that space is given by a kernel function $k_1(\mathbf{x}_i,\mathbf{x}_j)=\langle\phi(\mathbf{x}_i),\phi(\mathbf{x}_j)\rangle$. Similarly, $\mathcal{G}$ is the second kernel space on $\mathcal{L}$ with kernel function $k_2(\mathbf{y}_i,\mathbf{y}_j)=\langle\varphi(\mathbf{y}_i),\varphi(\mathbf{y}_j)\rangle$.
For a series of $N$ independent observations drawn from $p_{\mathbf{xy}}$, $\mathcal{Z}:=\{(\mathbf{x}_i,\mathbf{y}_i)\}_{i=1}^{N}$,  HSIC can be approximated as follows:
\begin{displaymath}
{\rm{HSIC}}(\mathbf{x}, \mathbf{y})=(N-1)^{-2}{\rm{tr}}(\mathbf{K}_1\mathbf{H}\mathbf{K}_2\mathbf{H})
\end{displaymath}
where $\mathbf{K}_1,\mathbf{K}_2$ are the Gram matrices with $\mathbf{K}_1(i,j)=k_1(\mathbf{x}_i,\mathbf{x}_j)$ and $\mathbf{K}_2(i,j)=k_2(\mathbf{y}_i,\mathbf{y}_j)$; $\mathbf{H}_{ij}=\delta_{ij}-1/N$ centers the Gram matrix to have zero mean, with $\delta_{ij}=1$ if $i=j$, and $\delta_{ij}=0$ otherwise.  Then, we can surrogate $\Omega(\cdot)$ with ${\rm{HSIC}}(\mathbf{D}_w, \mathbf{D}_v)$ and define  $\Omega(\cdot)$ as follows:
\begin{equation}\label{eq3}\small\vspace{-0.1cm}
  \Omega(\mathbf{D}_1, \cdots, \mathbf{D}_W)=\sum_{m=1}^M \sum_{w, v \in \mathcal{W}_m} - {\rm{HSIC}}(\mathbf{D}_w, \mathbf{D}_v)\vspace{-0.1cm}
\end{equation}
where $\mathcal{W}_m\subseteq\mathcal{W}$ includes the workers of the $m$-th group.

%\begin{equation}\label{eq3}
%  \Omega(\mathbf{D}^{(1)},\ldots,\mathbf{D}_w)=\sum_{w \ne v \atop w=1}^{|\mathcal{W}^*|} {\rm{IND}}(\mathbf{D}_w, \mathbf{D}_v)
%\end{equation}
%where $\mathbf{D}_w, \mathbf{D}_v$ are in the same community, $\mathcal{W}^*\subseteq\mathcal{W}$ denotes the number of workers in the same community. The aim of regularization ${\rm{IND}}(,)$ is to enhance the dependency  between different workers in $\mathcal{W}^*$.
Unlike single-label data, the labels of multi-labeled data are correlated. By properly exploring and leveraging label correlations we can boost  the learning performance, and also reduce the query cost \cite{chen2018cost}. To this end, we additionally guide the pursue of label correlations as follows:
\begin{equation}\label{eq4}\vspace{-0.1cm}
\small
\begin{split}
  &\Psi(\mathbf{C}_1, \ldots, \mathbf{C}_M)=\sum_{m=1}^M\frac{1}{2}\mathbf{B}_{ij}\|\mathbf{C}_m(i,\cdot)-\mathbf{C}_m(j,\cdot)
  \|^2_F \\ &=\sum_{m=1}^M{\rm{tr}}(\mathbf{C}_m^T(\mathbf{O}-\mathbf{B})\mathbf{C}_m)
  =\sum_{m=1}^M{\rm{tr}}(\mathbf{C}_m^T\mathbf{L}\mathbf{C}_m)\\
\end{split}\vspace{-0.1cm}
\end{equation}
$\mathbf{B}\in \mathbb{R}^{L\times L}$ stores the latent label correlations between $L$ labels.  $\mathbf{C}_m(i,\cdot)$ is the $i$-th row of $\mathbf{C}_m$, $\mathbf{O}$ is a diagonal matrix with $\mathbf{O}_{ii} = \sum_{k=1}^L \mathbf{B}_{ik}$, and $\mathbf{L} = \mathbf{O} - \mathbf{B}$. For simplicity, we adopt the widely used cosine similarity to quantify the latent correlations between labels, based on the averaged sample-label association matrix $\sum_{w=1}^W \mathbf {A}_w/W$. Other more advanced label correlation measurements can also be adopted  \cite{zhang2014review}.

To better learn the liner representation of each annotation matrix and groups, we combine the constraints on $\mathbf{D}_w$ and $\mathbf{C}_m$, and then formulate the multi-label consensus objective function of AMCC as follows:
\begin{equation}\label{eq5}\vspace{-0.1cm}
\small
\begin{aligned}
O=\mathop{\arg\min}\limits_{\mathbf{D}_w,\mathbf{C}_m}&\sum_{m=1}^M\sum_{w=1}^W\boldsymbol{\lambda}_m^r\|\mathbf{A}_w-\mathbf{A}_w
  (\mathbf{D}_w+\mathbf{C}_m)\|^2_F + \\
  &\alpha \sum_{m=1}^M \sum_{w, v \in \mathcal{W}_m} - {\rm{HSIC}}(\mathbf{D}_w, \mathbf{D}_v)
  +\beta\sum_{m=1}^M{\rm{tr}}(\mathbf{C}_m^T\mathbf{L}\mathbf{C}_m) \\
  &s.t. \ \ \boldsymbol{\lambda}_m \geq 0,\ \sum_{m=1}^M\boldsymbol{\lambda}_m=1, \ \sum_{l=1}^L\mathbf{D}_w(g,l)=1
\end{aligned}
\end{equation}
where the parameters $\alpha$ and $\beta$ weight the constraints in Eq. (\ref{eq3}) and Eq. (\ref{eq4}), respectively. Our experiments confirm the advantage of including these two constraints.

Each sample is often independently annotated by worker $w$, and the number of annotations of each worker  has a multinomial distribution \cite{dawid1979maximum}. The consensus labels should  account for the specialty $\mathbf{D}_w$ and the commonality ($\mathbf{C}_m$) of workers. As such, we can then compute the multi-label consensus labels as follows:
\begin{equation}\label{eq6}\small
\begin{split}
  \hat{\mathbf{y}}_i&=\mathop{\arg\max}\limits_{g\in\mathcal{L}}P(\mathbf{y}_i|\mathbf{A}_w, \mathbf{D}_w^*, \mathbf{C}_m^*)=\mathop{\arg\max}\limits_{g\in\mathcal{L}}\\
&\sum_{m=1}^M (\boldsymbol{\lambda}_m^*)^r \prod_{w=1}^W\prod_{l=1}^L (\mathbf{C}_m^*(g,l)+\mathbf{D}_w^*(g,l))^{\mathbb{I}(\mathbf{A}_w(i,l)=1)}
\end{split}
\end{equation}
where $\mathbf{D}_w^*$,  $\mathbf{C}_m^*$, and $\boldsymbol{\lambda}_m^*$ are the optimized (detail procedures are in Section \ref{sec:optimization}) values of Eq. (\ref{eq5}). $\hat{\mathbf{y}}_i$ is the consensus labels of sample $\mathbf{x}_i$, and $\mathbb{I}(\cdot)$ is the indicator function, which returns 1 if the argument is true and 0 otherwise.

\subsection{Cost-saving Active Crowdsourcing Learning}
\label{activelearning}
In practice, the budget for crowdsourcing is often limited. Annotating all the samples is infeasible and may lead to unnecessary information redundancy. As such, we further study AMCC in combination with active learning, to account for the different expertise and cost of workers. In each iteration of active learning, we select a cost-effective sample-label-worker triplet $(\mathbf{x}^*, l^*, w^*)$ with the following two properties: (i) the selected sample-label pair $(\mathbf{x}^*, l^*)$ is the \emph{most useful} for the improvement of AMCC; (ii) the selected worker $w^*$ can reliably annotate  the sample $\mathbf{x}^*$ with the label $l^*$ at the \emph{lowest possible cost}.

\subsubsection{Sample-label selection}
Quantifying how useful a sample is towards the improvement of a prediction model is the key task of active learning, and various quantifying criteria have proposed \cite{settles09}. In this work, we use uncertainty, which is widely used in the literature \cite{ye2015multi}. If the current model is uncertain about the prediction on a sample, gathering label information about that sample may provide useful knowledge which is not yet embedded in the model. We simply estimate the uncertainty of  a sample-label pair as follows:
\begin{equation}\label{eq7}\small
  u^1(i,l)=1-|\frac{1}{2}-p(\mathbf{y}_{il}=1|\mathbf{x}_i)|
\end{equation}
where $p(\mathbf{y}_{il}=1|\mathbf{x}_i)$ reflects the probability that $l$ is a positive label of $\mathbf{x}_i$: $p(\mathbf{y}_{il}=1|\mathbf{x}_i) \rightarrow 1$ indicates that the predictor assigns  the $l$-th label  to $\mathbf{x}_i$ with confidence. Similarly, when $p(\mathbf{y}_{il}=1|\mathbf{x}_i) \rightarrow 0$ the predictor is confident that the $l$-th label  does not belong to $\mathbf{x}_i$. As such, larger $u^1(i,l)$ values are an indication of higher uncertainty. {We admit other criteria of the sample-label pair can also be adopted here, which is not the main focus of this paper.}

Label correlations play an important role in saving the query cost. If the selected label $l$ is positively correlated with other potential labels of the same sample, then  querying and annotating the sample  with $l$ would also reduce the uncertainty of the other labels \cite{chen2018cost,ye2015multi}. We can use the already learned  label correlations and weights associated to different groups, and form the integrated label correlation as $\mathbf{C}=\frac{1}{M}\sum_{m=1}^M\boldsymbol{\lambda}_m^r \mathbf{C}_m$.
We then quantify the expected uncertainty reduction when $l$ is queried for $\mathbf{x}_i$ as follows:
\begin{equation}\label{eq8}\small\vspace{-0.1cm}
u^2(i,l)=\left\{ \begin{array}{l}
\frac{1}{|\bar{\mathcal{L}}_i|}\sum\limits_{k=1}^L|\mathbf{C}_{lk}|\times\mathbb{I}(\mathbf{y}_{ik}\in \bar{\mathcal{L}}_i), \ \ |\bar{\mathcal{L}}_i|>1
\\
0, \ \ |\bar{\mathcal{L}}_i|=1
\end{array} \right.
\end{equation}
where $\bar{\mathcal{L}}_i$ includes all the un-queried labels of $\mathbf{x}_i$. $u^2(i,l)$ averages the label correlations of the un-queried label $l$ with the other un-queried labels of the same example. A larger $u^2(i,l)$ value indicates that $l$ can reduce the uncertainty more than other un-queried labels.

Now, we can leverage the two uncertainty quantities $u^1$ and $u^2$ to measure the integrative uncertainty of the selected sample-label pair. The higher $u^1(i,l)$ is, the larger the uncertainty is. Similarly, the larger  $u^2(i,l)$ is, the larger the information gain is expected to be when $l$ is queried. Given this, we define the integrative uncertainty as follows:
\begin{equation}\label{eq9}\small\vspace{-0.1cm}
u(\mathbf{x}_i, l)= \eta u^2(i,l)+(1-\eta)u^1(i,l)
\end{equation}
where $\eta \in (0,1)$ is a scalar coefficient to balance the importance of these two uncertainties. {In this paper, we fix $\eta=0.3$ according to the experiments}. Based on the integrative uncertainty, we can select the most useful sample-label pair as follows:
\begin{equation}\label{eq10}\small\vspace{-0.1cm}
(\mathbf{x}^*,l^*)=\mathop{\arg\max}\limits_{\mathbf{x}\in \mathcal{D}, l\in \mathcal{L}} u(\mathbf{x}, l)
\end{equation}

\subsubsection{Worker selection}
Each worker has his/her own specialty. As a result, different workers are good at annotating different samples, and a worker with low overall quality and low cost may still give reliable annotations to specific samples. Therefore, it is not suitable to approximate the credibility of a worker on a specific sample using his overall annotation quality. Here, we assume that a worker's credibility in annotating $\mathbf{x}_i$ can be approximated based on his submitted annotations to neighbor samples ($\mathbf{x}_j \in \mathcal{D}^L$) of $\mathbf{x}_i$ and on the consensus annotations $\hat{\mathbf{y}}_j$ computed using Eq. (\ref{eq6}). Since the crowd workers are divided into $M$ groups, each group has its own annotation behaviour (i.e., bias and label correlations), which is captured in $\mathbf{C}_m$. In addition, each worker also has his bias towards the $L$ distinct labels, which is encoded in $\mathbf{D}_w$. Given this, we can estimate the credibility of worker $w$ towards annotating  $\mathbf{x}_i$ as follows:
\begin{equation}\label{eq11}
\small
\begin{split}
  q_w(\mathbf{x}_i)= \frac{1}{k}\sum_{\mathbf{x}_j\in \mathcal{N}_k(\mathbf{x}_i), \atop w\in \mathcal{W}^m} \mathbf{S}(\mathbf{x}_i,\mathbf{x}_j)P^w(\mathbf{a}^w_j=\hat{\mathbf{y}}_j|\mathbf{C}_m, \mathbf{D}_w, \mathbf{x}_j)\\
  {\rm{where}} \ \ P^w(\cdot)=\prod_{g=1}^L\prod_{l=1}^L(\mathbf{C}_m(g,l)+\mathbf{D}_w(g,l))^{\mathbb{I}(\mathbf{A}^w_{jl}=1)}
\end{split}
\end{equation}
where $\mathcal{N}_k(\mathbf{x}_i)$ includes the $k$ nearest neighbors  of $\mathbf{x}_i$ in $\mathcal{D}^L$; $\mathbf{S}(\mathbf{x}_i, \mathbf{x}_j)$ stores the similarity between $\mathbf{x}_i$ and $\mathbf{x}_j$, which is the inverse of their Euclidean distance. $P^w(\cdot)$ computes the probability that the $w$-th worker correctly annotates $\mathbf{x}_j$. Similar as Eq. (\ref{eq6}), the annotations have a multinomial distribution.
%which indicates that the worker who gives accurate annotations for most neighbors of $\mathbf{x}_j$ is more reliable.
Obviously, among the $k$ neighbors, samples more similar to $\mathbf{x}_i$ contribute more to the estimation of $q_w(\mathbf{x}_i)$.

A worker who provides high-quality annotations in crowdsourcing should be appropriately rewarded. Here, we can approximate the cost of a worker using the credibility of his previous annotations. Based on the specificity $\mathbf{D}_w$ of the $w$-th worker and his previous annotations, we can approximate his cost $c_w$ as follows:
\begin{equation}\label{eq12}\small
  c_w=reward(\frac{1}{nnz(\mathbf{A}_w)}\sum_{i=1}^{N}\prod_{l=1}^L \mathbf{D}_w(l,l)^{\mathbb{I}(\mathbf{a}_{il}^w = 1)})
\end{equation}
where $\frac{1}{nnz(\mathbf{A}_w)}\sum_{i=1}^{N}\prod_{l=1}^L \mathbf{D}_w(l,l)^{\mathbb{I}(\mathbf{a}_{il}^w = 1)}$ estimates the overall annotation quality of the worker, and $nnz(\cdot)$ counts the number of nonzero entries of $\mathbf{A}_w$, which is equal to the number of annotations provided by the $w$-th worker. $reward(\cdot)$ is a user-specified reward function. In this paper, for simplicity, the reward function used is  linear: $reward(x)=x$.

\subsubsection{Cost-effective sample-label-worker selection}
To achieve  cost-effective sample-label-worker triplets $(\mathbf{x}^*, l^*, w^*)$, we need a large integrative uncertainty $u(\mathbf{x}^*,l^*)$, a  high-credible worker $q_w(\mathbf{x}^*)$, and a cost $c_w$ as low as possible. A straightforward criterion for selecting the best sample-label-worker triplet $(\mathbf{x}^*, l^*, w^*)$ is:
\begin{equation}\label{eq13}\small\vspace{-0.1cm}
  (\mathbf{x}^*, l^*, w^*)=\mathop{\arg\max}\limits_{\mathbf{x}\in \mathcal{D}, l\in \mathcal{L}, w\in W}\frac{u(\mathbf{x},l)q_w(\mathbf{x})}{c_w}\vspace{-0.1cm}
\end{equation}
A sample-label-worker triplet that violates any of the three desired properties will receive a small score in Eq. (\ref{eq13}). We emphasize that \emph{none} of the existing cost-effective crowdsourcing solutions \cite{yan2011active,Fang2014Active} can jointly account for the impact of workers, labels, and samples in crowdsourcing. Our experimental results  confirm the benefits of this effort.

\subsection{Optimization}
\label{sec:optimization}
\subsubsection{Algorithm Optimization}
Inspired by the idea of the Alternating Direction Method of Multipliers \cite{boyd2004convex}, we adopt the alternative minimization strategy to solve Eq. (\ref{eq5}).

\textbf{Updating} $\mathbf{D}_w$ \textbf{with fixed} $\boldsymbol{\lambda}_m$, $\mathbf{C}_m$. We need to minimize the following objective function
\begin{equation}\label{eq15}
\vspace{-0.3cm}
\scriptsize
\begin{split}
  \mathcal{L}(\mathbf{D}_w)&=\sum_{m=1}^M\sum_{w=1}^W\boldsymbol{\lambda}_m^r\|\mathbf{A}_w-\mathbf{A}_w
  (\mathbf{D}_w+\mathbf{C}_m)\|^2_F  -\beta\sum_{w, v \in \mathcal{W}_m}{\rm{HSIC}}(\mathbf{D}_w, \mathbf{D}_v)\\
  & s.t. \ \ \ \|\mathbf{d}_{i.}^{(w)}\|_2=1
\end{split}
\vspace{-0.4cm}
\end{equation}
We introduce an auxiliary variable $\mathbf{S}^{(w)}$, and then obtain the following objective
\begin{equation}\label{eq16}
\scriptsize
\vspace{-0.3cm}
\begin{split}
  \mathcal{L}(\mathbf{D}_w)&=\sum_{m=1}^M\sum_{w=1}^W\boldsymbol{\lambda}_m^r\|\mathbf{A}_w-\mathbf{A}_w
  (\mathbf{D}_w+\mathbf{C}_m)\|^2_F-\beta\sum_{w, v \in \mathcal{W}_m}{\rm{HSIC}}(\mathbf{D}_w, \mathbf{D}_v)\\
  & s.t. \ \mathbf{D}_w=\mathbf{S}^{(w)}, \  \|\mathbf{s}_{i.}^{(w)}\|_2=1
\end{split}
\vspace{-0.3cm}
\end{equation}
 By removing the equality constraint, Eq. (\ref{eq16}) becomes
\begin{equation}\label{eq17}
\scriptsize \vspace{-0.3cm}
\begin{split}
  & \mathcal{L}(\mathbf{D}_w, \mathbf{S}^{(w)}, \mathbf{T}^{(w)} )=\sum_{m=1}^M\sum_{w=1}^W\boldsymbol{\lambda}_m^r\|\mathbf{A}_w-\mathbf{A}_w
  (\mathbf{D}_w+\mathbf{C}_m)\|^2_F \\ &-\beta\sum_{w \ne v \atop w=1}^W {\rm{HSIC}}(\mathbf{D}_w, \mathbf{D}_v)+\mu\|\mathbf{D}_w-\mathbf{S}_r^{(w)}+\mathbf{T}_r^{(w)}\|_F^2
  \\
  & s.t. \ \  \|\mathbf{s}_{i.}^{(w)}\|_2=1
\end{split}\vspace{-0.3cm}
\end{equation}
where $\mu > 0$ is the penalty hyperparameter. The optimal solution of Eq. (\ref{eq17}) can be obtained with
\begin{equation}\label{eq18}
\small\vspace{-0.1cm}
\left\{ \begin{array}{l}
\mathbf{D}_{r+1}^{(w)}=\arg\min\limits_{\mathbf{D}_{r}^{(w)}}
\sum\limits_{m=1}^M\|\mathbf{A}_w-\mathbf{A}_w
  (\mathbf{D}_w+\mathbf{C}_m)\|^2_F  \\
-\beta\sum\limits_{w \ne v \atop w=1}^{W^*} {\rm{HSIC}}(\mathbf{D}_w, \mathbf{D}_v)+\mu\|\mathbf{D}_w-\mathbf{S}_r^{(w)}+\mathbf{T}_r^{(w)}\|_F^2
\\
\mathbf{S}_{r+1}^{(w)}=\arg\min\limits_{\mathbf{S}_{r}^{(w)}}\mu\|\mathbf{D}_{r+1}^{(w)}-\mathbf{S}_r^{(w)}+\mathbf{T}_r^{(w)}\|_F^2, \ s.t. \ \|\mathbf{s}_{i.}^{(w)}\|_2=1
\\
\mathbf{T}_{r+1}^{(w)}=\mathbf{T}_{r}^{(w)}+\mathbf{D}_{r+1}^{(w)}-\mathbf{S}_{r+1}^{(w)}, {\rm{update}} \ \mu \ {\rm{if \ appropriate}}
\end{array} \right.
\end{equation}

\textbf{Updating} $\mathbf{C}_m$ \textbf{with fixed} $\{\boldsymbol{\lambda}_m\}_{m=1}^M$, $\mathbf{D}_w$. We need to minimize the following objective function
\begin{equation}\label{eq19}
\small
\begin{split}
  \mathcal{L}(\mathbf{C}_m)&=\sum_{w=1}^W\boldsymbol{\lambda}_m\|\mathbf{A}_w-\mathbf{A}_w
  (\mathbf{D}_w+\mathbf{C}_m)\|^2_F \\ &+\alpha\sum_{m=1}^M\boldsymbol{\lambda}_m{\rm{tr}}((\mathbf{C}_m)^T\mathbf {L}\mathbf{C}_m)\\
\end{split}
\end{equation}
By taking the derivative with respect to $\mathbf{C}_m$ and setting it to zero, we obtain
\begin{equation}\label{eq20}
\small
  \sum_{w=1}^W\boldsymbol{\lambda}_m((\mathbf{A}_w)^T\mathbf{A}_w\mathbf{C}_m-(\mathbf{A}_w)^T\mathbf{A}_w)+\alpha\boldsymbol{\lambda}_m\mathbf {L}\mathbf{C}_m=0
\end{equation}
Accordingly, we can update $\mathbf{C}_m$ with the following rule
\begin{equation}\label{eq21}
\small
  \mathbf{C}_m=\frac{\boldsymbol{\lambda}_m\sum\limits_{w=1}^W(\mathbf{A}_w)^T\mathbf{A}_w}{\boldsymbol{\lambda}_m\sum\limits_{w=1}^W(\mathbf{A}_w)^T\mathbf{A}_w+\alpha \boldsymbol{\lambda}_m\mathbf{L}}
\end{equation}

\textbf{Updating} $\boldsymbol{\lambda}$ \textbf{with fixed} $\mathbf{D}_w,\mathbf{C}_m.$
\begin{equation}\label{eq22}
\scriptsize
\begin{split}
  L(\boldsymbol{\lambda}, \gamma)=\sum\limits_{m=1}^M\sum\limits_{w=1}^W\boldsymbol{\lambda}_m^r\|\mathbf{A}_w-\mathbf{A}_w
  (\mathbf{D}_w+\mathbf{C}_m)\|^2_F-\gamma (\sum_{m=1}^M\boldsymbol{\lambda}_m-1)
\end{split}
\end{equation}
By setting to zero the derivative of Eq.\ref{eq22} with respect to $\boldsymbol{\lambda}$ and $\gamma$, we obtain the following updating rule
\begin{equation}\label{eq23}
\small
  \boldsymbol{\lambda}_m=\frac{\left({\sum\limits_{w=1}^W\|\mathbf{A}_w-\mathbf{A}_w
  (\mathbf{D}_w+\mathbf{C}_m)\|^2_F}\right)^{1/1-r}}{\sum\limits_{m=1}^M\left({\sum\limits_{w=1}^W\|\mathbf{A}_w-\mathbf{A}_w
  (\mathbf{D}_w+\mathbf{C}_m)\|^2_F}\right)^{1/1-r}}
\end{equation}
According to the above  rules, we can alternatively update these variables until the convergence condition (i.e., the difference of the objective function value between two consecutive iterations is smaller than $10^{-5}$) is reached.

\subsubsection{Convexity Analysis}
{Because of the HSIC term involved in Eq. (\ref{eq3}) and (\ref{eq5}), it is generally not convex due to the negative sign. Therefore, we must make sure that the function in Eq. (\ref{eq17}) is convex. Obviously, we could obtain  the optimal solution above if $\mathcal{L}(\mathbf{D}_w)$ in Eq. (\ref{eq17}) is strictly convex, which is also a prerequisite for the convergence of the holistic optimization. Therefore, we explore the suitable parameter setting and ensure the convexity of $\mathcal{L}(\mathbf{D}_w)$ as follows:
\begin{theorem}\label{the1}
Given the parameter setting $\mu/ \beta \geq 4L(W-1)$, the subproblem $\mathcal{L}(\mathbf{D}_w)$ is convex, where $L,W$ are the number of labels and workers, respectively.
\end{theorem}
\begin{IEEEproof}
From \cite{boyd2004convex}, whether the Hessian matrix $\nabla^2\mathcal{L}(\mathbf{D}_w)$ is semi-positive definite or not decides the convexity of $\mathcal{L}(\mathbf{D}_w)$. We find that the first term $\sum_{m=1}^M\sum_{w=1}^W\boldsymbol{\lambda}_m^r\|\mathbf{A}_w-\mathbf{A}_w
  (\mathbf{D}_w+\mathbf{C}_m)\|^2_F$ is convex.  As a result, we only must ensure the strict convexity for the last two terms, as follows:
\begin{displaymath}
\mathcal{L}\left(\mathbf{D}_{w}\right)=-\beta \sum_{w \neq v \atop w=1}^{W} \operatorname{HSIC}\left(\mathbf{D}_{w}, \mathbf{D}_{v}\right)+\mu\left\|\mathbf{D}_{w}-\mathbf{S}_{r}^{(w)}+\mathbf{T}_{r}^{(w)}\right\|_{F}^{2}
\end{displaymath}
Fortunately, we can easily compute the Hessian matrix $\nabla^2\mathcal{L}(\mathbf{D}_w)$ as:
\begin{displaymath}
\nabla^{2} \mathcal{L}\left(\mathbf{D}_{w}\right)=\mu \mathbf{I}-\beta \sum_{u \neq v}^{W} \mathbf{C D}_{w}^{T} \mathbf{D}_{w} \mathbf{C}=\mu \mathbf{I}-\beta \mathbf{Q}=\mathbf{P}
\end{displaymath}
For convenience, we let $ \sum_{u \neq v}^{W} \mathbf{C D}_{w}^{T} \mathbf{D}_{w} \mathbf{C} = \mathbf{Q}$. According to the Gerschgorin theorem \cite{Varga1962Matrix},  all the eigenvalues $\xi$ of $\mathbf{P}$ lie in $\left|\xi-\mu / \beta-\mathbf{Q}_{i i}\right| \leq \sum_{j \neq i}^{L}\left|\mathbf{Q}_{i j}\right|$. The value of $\mu / \beta$ will  satisfy the following constraint  after transformation:
\begin{displaymath}
\mu / \beta \geq \max _{1 \leq i \leq L}\left\{\sum_{j \neq i}^{L}\left|\mathbf{Q}_{i j}\right|-\mathbf{Q}_{i i}\right\}
\end{displaymath}
From the equation above, we can easily obtain  $\left|\mathbf{Q}_{i j}\right| \leq 4|W-1|$, therefore, the lower bound of $\mu / \beta$ is $4L(W -1)$. Accordingly, we can set $\mu =4L(W-1)\beta$ or even larger to ensure the constraint satisfied in practice.
\end{IEEEproof}

It comes to a conclusion that Theorem \ref{the1} guarantees the convexity of $\mathcal{L}(\mathbf{D}_w)$ and the subsequent optimal solution.}

\subsubsection{Complexity Analysis}
For multi-label crowd consensus, there are three  main sub-problems ($\mathbf{C}_m$, $\mathbf{D}_w$ and $\boldsymbol{\lambda}_m$) in our optimization procedure. Solving them has the following costs:  $O(t_1 WN_lL^2)$, $O(t_1 ML^3)$, and $O(t_1 MN_lL^2)$, respectively.  $t_1$ is the number of iterations in each consensus round (100 in this paper). For active learning, it takes $O(MLN_u)$, $O(kN_lL^2+N_l^2)$, and $O(nnz(\mathbf{A}_w)L)$ to select sample-label pairs, workers, and costs, respectively.
So the overall time complexity is $O(t_2 (MLN_u+kN_lL^2+N_l^2+nnz(\mathbf{A}_w)L))$, where  $t_2$ is the number of queries. {In practice, AMCC takes about 18 minutes on the SONYC-UST dataset and at most 4 minutes on the other six real-world datasets in Table \ref{dataset}, and 7 minutes  on the three simulated datasets in Table \ref{Table3} on a moderate PC.}

\section{Experimental Results and Analysis}
\label{exp}
\subsection{Experimental setup}
\textbf{Datasets:} To study the performance of AMCC in computing crowd consensus of multi-label samples, we perform experiments on {seven} real-world datasets. The statistics of the datasets are listed in Table ~\ref{dataset}. {\emph{SONYC Urban Sound Tagging (SONYC-UST)} \footnote{https://zenodo.org/record/3233082\#.XSXRhI} is a dataset for the development and evaluation of machine listening systems for realistic urban noise monitoring. We remove the workers who only annotated once, since his expertise is difficult to estimate. } \emph{Movie} is a movie category classification dataset used in \cite{Yoshimura2017Quality}.  \emph{Affective}\cite{snow2008cheap} includes a 100-headline sample  with six emotions, which collected annotations from Amazon Mechanical Turk. The workers were asked to provide scores between 0 and 100 for each emotion, with 0 meaning `not at all', and 100 meaning `maximum emotion'. The other four real-world datasets were used in emotion classification \cite{Duan2014Separate}. Ground truths are also provided for evaluating the consensus models.

\begin{table}[h!tbp]
\scriptsize
  \caption{Statistics of six real-world datasets used in the experiments}
  \begin{tabular}{lcccccc}
    \toprule
    \bfseries Datasets &
    \bfseries WRK. &
    \bfseries INS. &
    \bfseries LAB.  &
    \bfseries ANN. &
    \bfseries WpI. &
    \bfseries LpI. \\
    \midrule
    SONYC-UST &118 &442 &23 &1330 &3 &2.98\\
    Movie & 89  & 100 & 19 &6811 & 35 & 1.95\\
    Affective &38 &100 &6 &6000&10 &6\\
    AppleEkman & 68  & 78  & 6  &2978 & 30 & 1.27\\
    AppleNakamura & 57  & 78  & 10  &2768 & 30 & 1.18\\
    LoveEkman & 54  & 63  & 6  &1890  & 30 & 1.05\\
    LoveNakamura & 41  & 63  & 10  &3965 & 41 & 1.53\\

    \midrule
    \multicolumn{7}{l}{
    \textbf{WRK.}: number of workers; \textbf{INS.}: number of instances; }\\
    \multicolumn{7}{l}{\textbf{LAB.}: number of labels; \textbf{ANN.}: number of annotations; }\\ \multicolumn{7}{l}{\textbf{WpI.} (\textbf{LpI.}): Average number of workers (labels) per Instance.} \\
    \bottomrule
    \end{tabular}%
\label{dataset}
\end{table}%

\textbf{Comparing methods:} To perform a comparative study of AMCC, we conduct two types of experiments. Our first goal is to compare the performance of AMCC on the task of computing the crowd consensus against MV \cite{Sheng2008Get} , C-DS \cite{Duan2015Crowdsourced}, RA$k$EL-GLAD \cite{Yoshimura2017Quality}, MCMLD \cite{zhang2018multi}, ML-JMF \cite{Tu2018multi}, and NAM \cite{li2018multi}. The first two are classical single-label crowdsourcing solutions, and the other four are multi-label crowd consensus solutions.

Our second goal is to compare the active learning strategy of AMCC against its variants AMCC(rW), AMCC(rSL), AMCC(nLC), and against other baselines, which include {SAC \cite{yan2011active}}, MV(rWS), and NAM(A). AMCC(rW) randomly selects workers for active learning; AMCC(rSL) randomly selects the sample-label pairs to be queried; AMCC(nLC) does not use  label correlations, and it only uses Eq. (\ref{eq8}) to select sample-label pairs.  SAC is a single-label active crowdsourcing solution; for the query, it selects the most uncertain sample from the most reliable worker (with the highest accuracy)\cite{yan2011active}; NAM(A) is an active multi-label crowdsourcing method, it uses QCI (Query label Cardinality Inconsistency) \cite{Huang2013Active} to select the samples, and selects the most reliable worker for query; MV(rSW) randomly selects a sample and a set of workers for the query, and then uses the majority vote rule to generate the label for the queried sample.

We implemented MCMLD based on its original paper, and adopted the original codes for the other methods. The input parameters of the baseline methods are set and/or optimized as recommended by the authors.  {AMCC and its variants achieve a good performance when the scalar parameters $\alpha=0.1$, $\beta=10$, $r=2$ and $M=5$ in Eq. (\ref{eq2}). The parameter sensitivity analysis of them will be reported. }

\textbf{Evaluation metrics:}
Three widely used metrics are adopted for performance comparisons: Accuracy, Ranking Loss (RL),
and OneError (OE).  In multi-label crowd consensus, results can be partially correct. We therefore rely on the set-based definition of Accuracy to evaluate the individual correctness on $N$ samples \cite{Tu2018multi,hung2018computing}:
\begin{equation}\small
Accuracy=\frac{1}{N} \sum_{i=1}^N \frac{|\mathcal{T}_i \bigcap \mathcal{T}_i^*|}{|\mathcal{T}^*|}
\end{equation}
where $\mathcal{T}_i$ and $\mathcal{T}_i^*$ are the sets of true labels and the set of consensus labels of the $i$-th sample, respectively. For consistency with the other evaluation metrics, we report 1-RL (1-OE) instead of RL (OE). Thus, like Accuracy, the higher the value of 1-RL (1-OE) is, the better the performance will be. These metrics evaluate multi-label learning from different perspectives; as such, it's expected that a single method may perform well on some, but not necessarily on all metrics. Formal definitions of the metrics can be found in \cite{zhang2014review}.

\subsection{Consensus annotation results}
Here we present and discuss the first set of experiments. For a fair comparison, AMCC does {\textit{not}} include the active learning process of sample-label-worker triplets for the query.  We independently run each  method ten times and report the average results.
\subsubsection{Results on simulated data}
To cope with the lack of real-world multi-label crowdsourcing datasets that cover a variety of scenarios, we also conduct   experiments on simulated datasets, which can further reveal the difference among the compared algorithms under various conditions.

We selected three widely used multi-label datasets for classification from  MULAN\footnote{http://mulan.sourceforge.net/datasets-mlc.html}, and give the details in Table \ref{Table3}. The labels of these datasets naturally exhibit correlations. We observe that, rather than carefully annotating samples with each appropriate label, the crowd worker prefers to scan and annotate samples with the most relevant labels, and leave the remaining labels untouched. We simulate this behavior as follows.  We randomly bisect each dataset into two parts, denoted as DATA1 and DATA2. We use the classic multi-label classifier RankSVM \cite{elisseeff2002kernel} to predict the relevance rank of labels to simulate the annotation process. RankSVM is separately trained on \{90\%, 80\%, 70\%, 60\%, 50\%, 40\%, 30\%\} of the samples of  DATA1, thus simulating seven workers. Then, we use the trained RankSVM to make prediction on  80\% of the instances randomly selected from DATA2, and take the predicted results as the annotations provided by a worker. In addition, we add six `workers' (spammers): three uniformly annotate all samples with a specific label; the other three separately annotate a sample with a random label. Table \ref{Table4} shows the results of the  methods on the simulated datasets.
\begin{table}[h!tbp]
\scriptsize
%\scriptsize
  \centering
  \caption{Three multi-label datasets for simulations.}\vspace{-0.3cm}
  \begin{tabular}{ccccccc}
  \\
 \toprule
 \bfseries Dataset  & \bfseries INS. & \bfseries LAB.  & \bfseries LpI.   & \bfseries FEA. & \bfseries  ANN.\\
 \midrule
 Emotions & 593 & 6 & 1.870 &6 & 1646 \\
 Scene & 2407 & 6 & 1.074 &6 & 6742\\
 Yeast & 2417 & 14 & 4.237  &14 & 6770 \\
 \midrule
 \multicolumn{6}{l}{
 \textbf{FEA}: number of features}\\
 \bottomrule
 \end{tabular}\label{Table3}
\end{table}%

\begin{table}[ht!bp]
\scriptsize
    \caption{Average results and standard deviations of AMCC and comparing methods on simulated datasets. $\circ/\bullet$ indicates AMCC is statistically  worse/better than the comparing method, and the statistical significance is assessed using a pairwise $t$-test at 95\% confidence level.}
     \begin{center}
    \begin{tabular}{c|c c c}
    \toprule
     {Methods/Metrics} &\textbf{Accuracy} & \textbf{1-RL} & \textbf{1-OE}\\
    \hline
    & \multicolumn{3}{c}{Yeast}\\
     \cline{2-4}
    \textbf{MV} &$0.815\pm0.000\bullet$   &$0.532\pm0.000\bullet$    &$0.567\pm0.000\bullet$  \\
    \textbf{C-DS}   &$0.832\pm0.020\bullet$  &$0.652\pm0.014\bullet$  &$0.556\pm0.024\bullet$   \\
    \textbf{RA$k$EL-GLAD} &$0.855\pm0.013\bullet$ &$0.646\pm0.013\bullet$ &$0.652\pm0.014\bullet$ \\
    \textbf{MCMLD}  &$0.871\pm0.013\bullet$   &$0.656\pm0.023\bullet$  &$0.632\pm0.024\bullet$ \\
    \textbf{ML-JMF}  &$0.881\pm0.012\bullet$   &$0.666\pm0.016\bullet$  &$0.662\pm0.016\bullet$ \\
    \textbf{NAM}  &$0.865\pm0.008\bullet$   &$0.726\pm0.000\bullet$  &$0.771\pm0.010\circ$ \\
    \textbf{AMCC}  &$0.892\pm0.003$   &$0.741\pm0.003$  &$0.769\pm0.022$ \\
    \hline
    & \multicolumn{3}{c}{Emotions}\\
     \cline{2-4}
     \textbf{MV} &$0.725\pm0.000\bullet$   &$0.592\pm0.000\bullet$    &$0.412\pm0.000\bullet$  \\
    \textbf{C-DS}   &$0.762\pm0.012\bullet$  &$0.642\pm0.014\bullet$  &$0.486\pm0.014\bullet$   \\
    \textbf{RA$k$EL-GLAD} &$0.795\pm0.011\bullet$ &$0.616\pm0.043\bullet$ &$0.421\pm0.044\bullet$ \\
    \textbf{MCMLD}  &$0.823\pm0.008\bullet$   &$0.646\pm0.033\bullet$  &$0.532\pm0.015\bullet$ \\
    \textbf{ML-JMF}  &$0.831\pm0.022\bullet$   &$0.687\pm0.028\bullet$  &$0.442\pm0.032\bullet$ \\
    \textbf{NAM}  &$0.827\pm0.010\bullet$   &$0.736\pm0.027\bullet$  &$0.542\pm0.009\bullet$ \\
    \textbf{AMCC}  &$0.842\pm0.013$   &$0.756\pm0.017$  &$0.564\pm0.011$ \\
    \hline
    & \multicolumn{3}{c}{Scene}\\
     \cline{2-4}
     \textbf{MV} &$0.705\pm0.000\bullet$   &$0.432\pm0.000\bullet$    &$0.456\pm0.000\bullet$  \\
     \textbf{C-DS}   &$0.762\pm0.012\bullet$  &$0.492\pm0.014\bullet$  &$0.486\pm0.014\bullet$ \\
    \textbf{RA$k$EL-GLAD} &$0.734\pm0.000\bullet$ &$0.522\pm0.023\bullet$ &$0.472\pm0.014\bullet$ \\
    \textbf{MCMLD}  &$0.751\pm0.000\bullet$   &$0.563\pm0.023\bullet$  &$0.556\pm0.014\bullet$ \\
    \textbf{ML-JMF}  &$0.749\pm0.022\bullet$   &$0.553\pm0.023\bullet$  &$0.563\pm0.014\bullet$ \\
    \textbf{NAM}  &$0.744\pm0.018\bullet$   &$0.617\pm0.023\circ$  &$0.572\pm0.014\bullet$ \\
    \textbf{AMCC}  &$0.771\pm0.012$   &$0.582\pm0.023$  &$0.592\pm0.014$ \\
    \bottomrule
    \end{tabular}
    \end{center}
\label{Table4}
\end{table}

We clearly see that AMCC frequently outperforms the comparing methods across different datasets and evaluation metrics. Besides Accuracy, AMCC generally has higher 1-RL values than other methods, which shows that AMCC can more reliably rank relevant labels ahead of irrelevant ones. The prominent results of AMCC on 1-OE again confirm this advantage.
RA$k$EL-GLAD, MCMLD, NAM, ML-JMF, and AMCC make use of label correlations and achieve better results than C-DS and MV, which do not use label correlations. MCMLD, NAM, ML-JMF, and AMCC achieve, most of the times, a better performance than RA$k$EL-GLAD and C-DS. This is because the latter two do not account for the quality variance of workers. MCMLD and NAM often lose to AMCC and ML-JMF, since they do not account for  workers' behaviors, whereas the latter two do. Both AMCC and ML-JMF can reduce the impact of spammers, but AMCC still achieves better results than ML-JMF. That is because AMCC models the workers by groups, and reduces the impact of sparse annotations by merging workers' annotations within the same group, and assigns weights to workers at the group level. In contrast, ML-JMF separately assigns a weight to each worker, and thus is more sensitive to annotation sparsity. As a result, AMCC can model the expertise of workers more reliably than ML-JMF. This observation supports our approach of separately accounting for the expertise of workers ($\mathbf{D}_w$) and for their annotation behaviors ($\mathbf{C}_m$) when computing crowd consensus labels \cite{kurve2015multicategory}.

\begin{figure}[h!tbp]
  \centering
  \includegraphics[width=7.5cm, height=4.65cm]{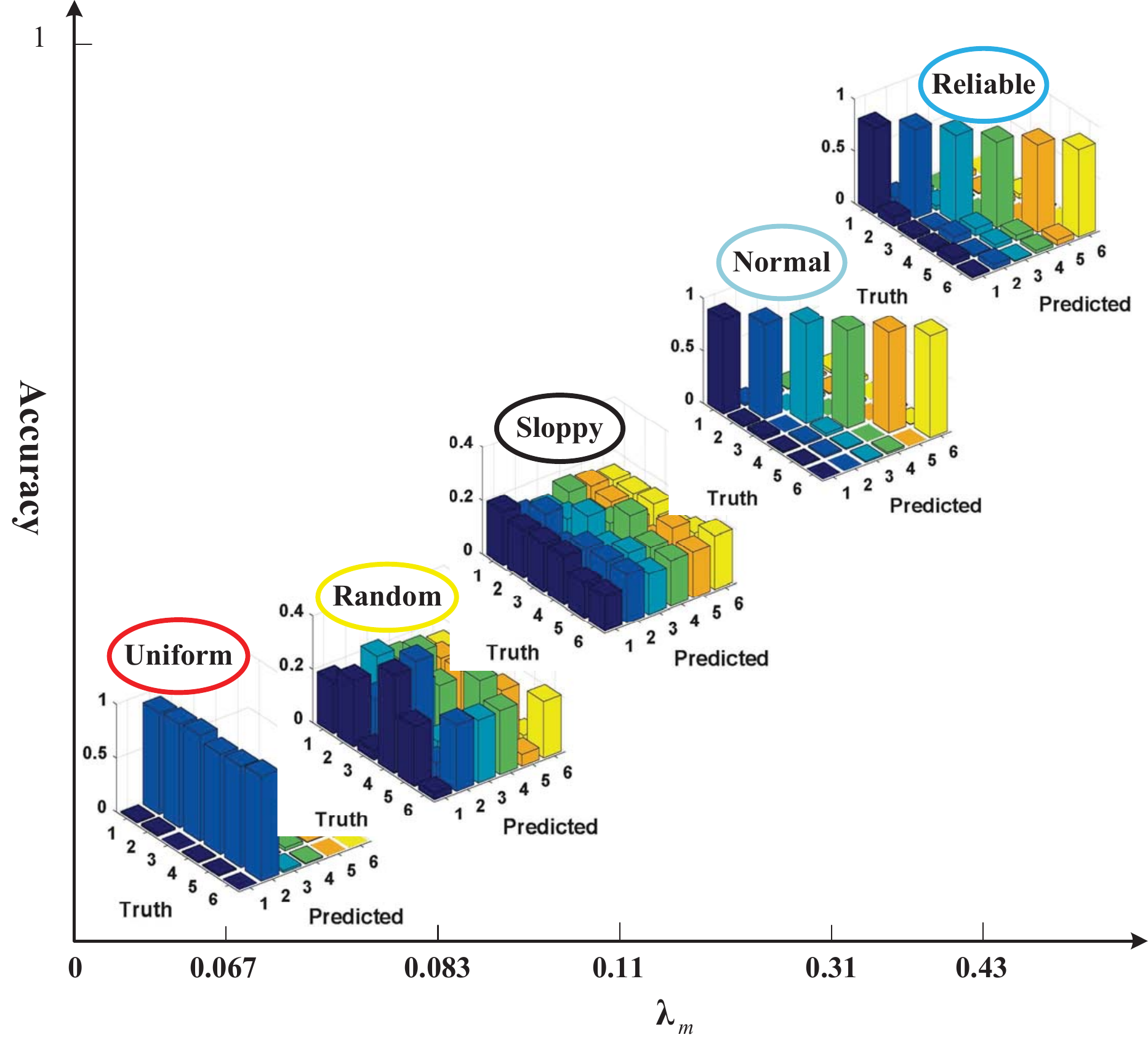}\\
  \caption{Accuracy, workers' individuality, and weights of groups on the Yeast dataset.}\label{type}
\end{figure}

Figure \ref{type} shows the results of AMCC in terms of worker individuality at the group level on the Yeast dataset. In the Figure, each set of 3D bars depicts the workers' individuality ($\mathbf{D}_w$), and the assigned weight $\boldsymbol{\lambda}_m$ of the group these workers belong to. We can see that the higher the weight $\boldsymbol{\lambda}_m$ is, the more reliable the workers in the corresponding group are. Reliable workers give correct annotations with high probability (diagonal values close to 1) and rarely provide wrong annotations. In the normal group, workers have relatively high accuracy, but the accuracy is smaller than that of the reliable group. Workers in the sloppy group often mistake a correct label with another label, which means that they often make incorrect annotations. AMCC assigns the lowest weight to the group of spammers, who randomly or uniformly annotate samples with labels and have the lowest accuracy.

In summary, our experimental results not only prove the effectiveness of AMCC in computing multi-label crowd consensus labels of samples, but also confirm that label correlations, the individuality, and the commonality of workers should be jointly leveraged. In addition, the results  justify the modelling of the specificity and of the commonality of workers at a group level, since doing so  reduces the number of weights and the impact of sparse annotations.
\subsubsection{Results on real-world datasets}
To evaluate the performance of AMCC in a real-world setting, we conduct experiments on the datasets listed in Table \ref{dataset}. The results  are reported in Table \ref{metric-result}.

\begin{table*}[ht!bp]
\scriptsize
    \caption{Average results and standard deviations on real datasets.
    $\circ/\bullet$ indicates AMCC is statistically  worse/better than the comparing method, and the statistical significance is assessed using a pairwise $t$-test at 95\% confidence level.}
     \begin{center}
    \begin{tabular}{c|c c c c c c |c}
    \toprule
     {Metrics} &\textbf{MV} &\textbf{C-DS} &\textbf{RA$k$EL-GLAD} &\textbf{MCMLD} &\textbf{ML-JMF}  &\textbf{NAM} &\textbf{AMCC}\\
    \hline
    & \multicolumn{7}{c}{Movie}\\
     \cline{2-8}
    Accuracy &$0.928\pm0.000\bullet$   &$0.942\pm0.001\bullet$    &$0.943\pm0.014\bullet$ &$0.951\pm0.003\bullet$ &$0.946\pm0.003\bullet$ &$0.944\pm0.010\bullet$ &$0.960\pm0.013$ \\
    \emph{1-RL}   &$0.934\pm0.000\bullet$     &$0.982\pm0.014\bullet$    &$0.986\pm0.024\bullet$   &$0.967\pm0.000\bullet$   &$0.987\pm0.000\bullet$ &$0.982\pm0.011\bullet$ &$0.988\pm0.009$\\
    \emph{1-OE}  &$0.894\pm0.000\bullet$   &$0.986\pm0.023\bullet$  &$0.982\pm0.014\bullet$ &$0.988\pm0.011\bullet$ &$0.987\pm0.021\bullet$ &$0.992\pm0.001\circ$ &$0.988\pm0.003$\\
    \hline
    & \multicolumn{7}{c}{AppleNakamura} \\
     \cline{2-8}
    Accuracy &$0.851\pm0.000\bullet$   &$0.928\pm0.001\bullet$    &$0.932\pm0.015\bullet$   &$0.937\pm0.006\bullet$ &$0.953\pm0.021\bullet$ &$0.945\pm0.010\bullet$ &$0.960\pm0.030$ \\
    \emph{1-RL}   &$0.856\pm0.000$   &$0.969\pm0.011\bullet$    &$0.968\pm0.017\bullet$ &$0.966\pm0.014\bullet$   &$0.970\pm0.011\bullet$ &$0.974\pm0.003\bullet$
    &$0.979\pm0.003$ \\
    \emph{1-OE}  &$0.868\pm0.000$   &$0.919\pm0.021\bullet$  &$0.975\pm0.022\bullet$ &$0.978\pm0.001\bullet$ &$0.979\pm0.000\circ$ &$0.981\pm0.003\circ$ &$0.978\pm0.000$\\
    \hline
    & \multicolumn{7}{c}{AppleEkman} \\
     \cline{2-8}
    Accuracy &$0.862\pm0.000\bullet$   &$0.936\pm0.014\bullet$    &$0.930\pm0.022\bullet$   &$0.904\pm0.003\bullet$ &$0.951\pm0.021\bullet$ &$0.943\pm0.000\bullet$ &$0.955\pm0.033$ \\
    \emph{1-RL}   &$0.901\pm0.000\bullet$   &$0.969\pm0.021\bullet$    &$0.970\pm0.016\bullet$ &$0.953\pm0.01\bullet$   &$0.972\pm0.017\bullet$ &$0.977\pm0.012\bullet$ &$0.982\pm0.012$ \\
    \emph{1-OE}  &$0.837\pm0.000\bullet$   &$0.869\pm0.017\bullet$  &$0.987\pm0.011\circ$ &$0.938\pm0.012\bullet$ &$0.989\pm0.001\circ $ &$0.970\pm0.002\bullet$ &$0.975\pm0.005$ \\
    \hline
    & \multicolumn{7}{c}{LoveNakamura} \\
     \cline{2-8}
    Accuracy &$0.873\pm0.000\bullet$   &$0.927\pm0.001\bullet$    &$0.936\pm0.021\bullet$   &$0.942\pm0.002\bullet$ &$0.951\pm0.013\circ$ &$0.942\pm0.007\bullet$ &$0.947\pm0.013$ \\
    \emph{1-RL}   &$0.933\pm0.000\bullet$   &$0.973\pm0.014\bullet$    &$0.991\pm0.021\circ$ &$0.976\pm0.000\bullet$   &$0.976\pm0.017\bullet$ &$0.981\pm0.014\circ$ &$0.979\pm0.000$ \\
    \emph{1-OE}  &$0.926\pm0.000\bullet$   &$0.988\pm0.021\bullet$  &$1.000\pm0.022\circ$ &$0.988\pm0.011\bullet$ &$0.989\pm0.024\bullet$ &$0.988\pm0.023\bullet$ &$0.991\pm0.004$ \\
    \hline
    & \multicolumn{7}{c}{LoveEkman}\\
     \cline{2-8}
    Accuracy &$0.870\pm0.000\bullet$   &$0.902\pm0.001\bullet$    &$0.920\pm0.014\bullet$ &$0.882\pm0.000\bullet$ &$0.923\pm0.012\bullet$ &$0.942\pm0.002\circ$ &$0.938\pm0.002$ \\
    \emph{1-RL}   &$0.563\pm0.000\bullet $     &$0.725\pm0.014\bullet$    &$0.817\pm0.024\bullet$   &$0.919\pm0.017\bullet$   &$0.936\pm0.000\bullet$ &$0.923\pm0.023\bullet$ &$0.943\pm0.031$ \\
    \emph{1-OE}  &$0.589\pm0.000\bullet$   &$0.787\pm0.023\bullet$  &$0.975\pm0.014\bullet$ &$0.957\pm0.011\bullet$ &$0.978\pm0.021\circ$ &$0.966\pm0.000\bullet$ &$0.977\pm0.011$ \\
    \hline
    &
    \multicolumn{7}{c}{Affective} \\
     \cline{2-8}
    Accuracy &$0.701\pm0.000\bullet$   &$0.727\pm0.001\bullet$    &$0.722\pm0.021\bullet$   &$0.738\pm0.002\bullet$ &$0.732\pm0.013\bullet$ &$0.737\pm0.005\bullet$ &$0.751\pm0.013$ \\
    \emph{1-RL}   &$0.562\pm0.000\bullet$   &$0.732\pm0.014\bullet$    &$0.787\pm0.021\bullet$ &$0.804\pm0.000\bullet$   &$0.812\pm0.017\bullet$ &$0.811\pm0.002\bullet$ &$0.822\pm0.022$ \\
    \emph{1-OE}  &$0.766\pm0.000\bullet$   &$0.862\pm0.021\bullet$  &$0.865\pm0.022\bullet$ &$0.865\pm0.011\bullet$ &$0.864\pm0.024\bullet$ &$0.874\pm0.005\circ$ &$0.870\pm0.023$ \\
    \hline
    &
    \multicolumn{7}{c}{SONYC-UST} \\
     \cline{2-8}
    Accuracy &$0.722\pm0.000\bullet$   &$0.810\pm0.001\bullet$    &$0.819\pm0.021\bullet$   &$0.805\pm0.002\bullet$ &$0.855\pm0.010\bullet$ &$0.843\pm0.005\bullet$ &$0.875\pm0.009$ \\
    \emph{1-RL}   &$0.662\pm0.000\bullet$   &$0.682\pm0.010\bullet$    &$0.697\pm0.011\bullet$ &$0.745\pm0.000\bullet$   &$0.701\pm0.007\bullet$ &$0.722\pm0.009\bullet$ &$0.765\pm0.010$ \\
    \emph{1-OE}  &$0.623\pm0.000\bullet$   &$0.613\pm0.011\bullet$  &$0.635\pm0.014\bullet$ &$0.667\pm0.011\bullet$ &$0.644\pm0.014\bullet$ &$0.662\pm0.005\bullet$ &$0.671\pm0.013$ \\
    \bottomrule
    \end{tabular}
    \end{center}
    \label{metric-result}
\end{table*}

\begin{figure}[h!tbp]
  \centering
  \includegraphics[width=9cm, height=5cm]{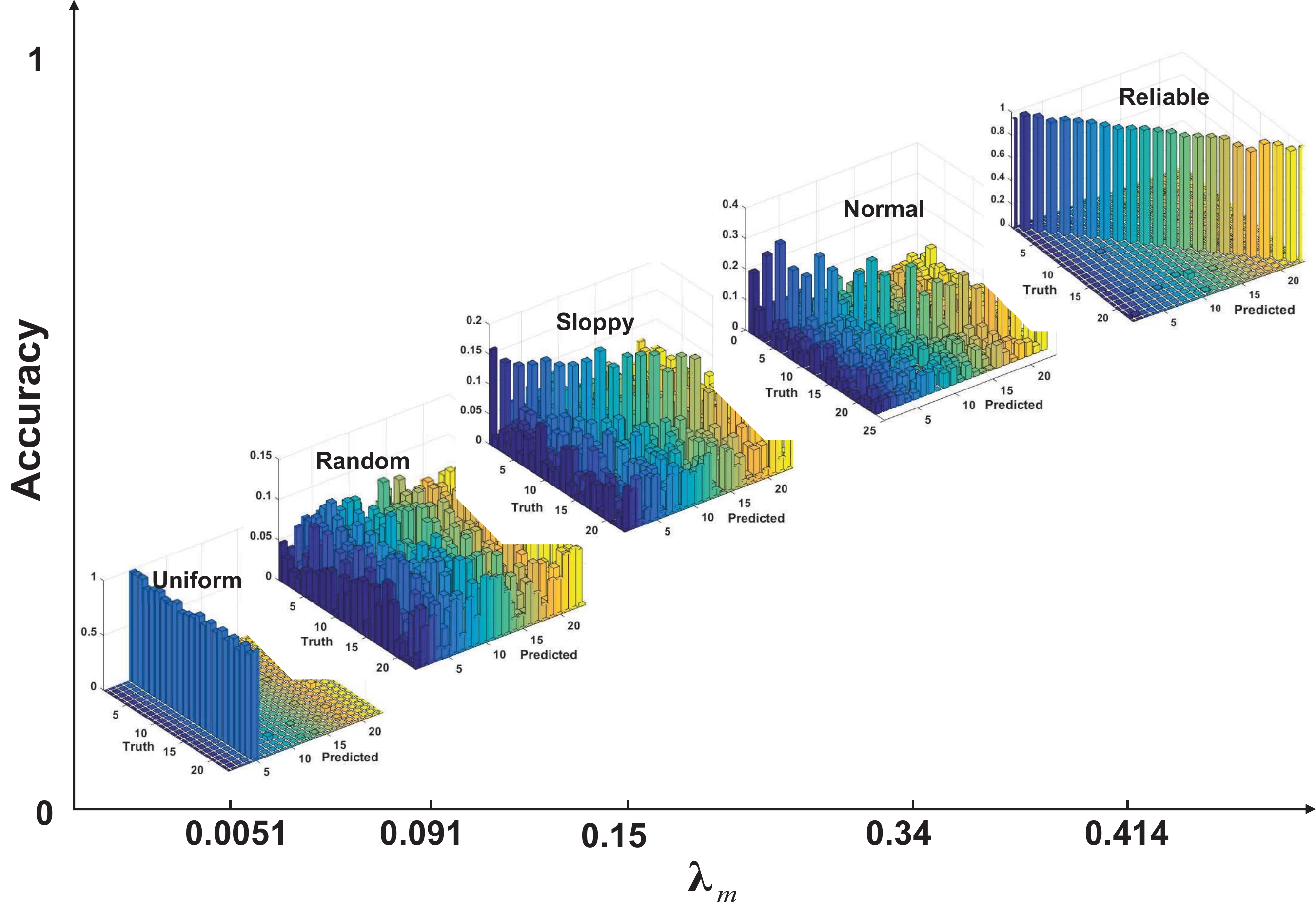}\\
  \caption{Accuracy, workers' group commonality ($\mathbf{C}_m$), and weights of groups on the SONYC-UST dataset.}\label{worker-sonyc}
\end{figure}

From Table \ref{metric-result} we can see that AMCC  achieves the highest accuracy in most of the cases. Specifically, the Accuracy  of AMCC on Movie and AppleNakamura  has obviously improved than others. RA$k$EL-GLAD, MCMLD, NAM, ML-JMF, and AMCC  all consider label correlations, and achieve higher Accuracy than the methods which do not account for label correlations. Furthermore, AMCC  generally has higher 1-RL and 1-OE values than C-DS, RA$k$EL-GLAD, MCMLD, ML-JMF, and NAM. {We also reveal $\mathbf{C}_m$ of AMCC on the SONYC-UST dataset in Figure \ref{worker-sonyc}. Alike Figure \ref{type}, AMCC clearly clusters workers into five different groups, and assigns different weights ($\boldsymbol{\lambda}_m$) to these groups. The lower the weight is, the less reliable the group (and workers within) is.}

\subsection{Active crowdsourcing results}
\label{active-result}
\begin{figure*}[h!tbp]
  \centering
  \includegraphics[scale=0.35]{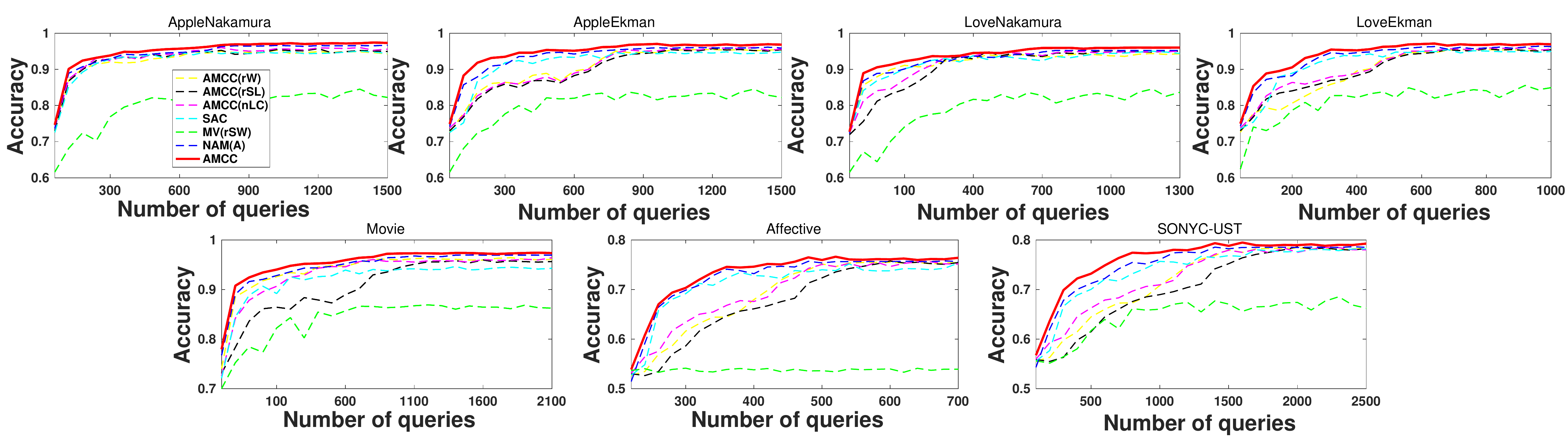}
  \caption{Accuracy of AMCC and comparing methods vs. number of queries on seven real datasets. }\label{active}
\end{figure*}

\begin{figure*}[h!tbp]
  \centering
  \includegraphics[scale=0.35]{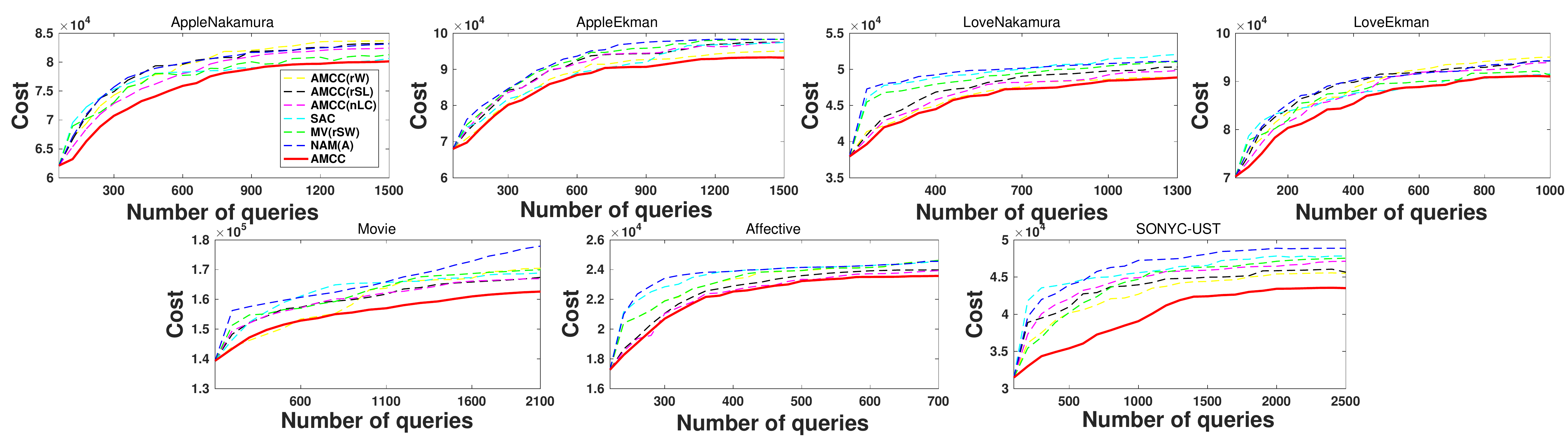}
  \caption{Cost of AMCC and comparing methods vs. number of queries  on seven real datasets.}\label{cost}
\end{figure*}

In this subsection, we conduct the \emph{second} type of experiments to study the effectiveness of the active crowdsourcing learning strategy on the {seven} real-world datasets. For each dataset, we randomly partition the samples into three parts, and separately use 5\%, 70\% and 25\% of the whole dataset to construct the initial labeled training data, the unlabeled training data, and the test data. We evaluate our consensus model in the active learning setting. For the cold-start case, the consensus outputs derived from initial worker annotations can serve as the labels of $\mathbf{D}^L$ and kick off the active learning process. We estimate the average accuracy of workers on the initial labeled set, we then set the worker's query cost within  one (lowest) and $W$ (highest), and proportionally to his quality. {We sample a batch of 5 instances at each iteration and repeat the iteration for 20 times.} After each iteration, the annotations returned by the selected sample-label-worker triplets are appended into $D^L$ and to update the learning model. The average performance over ten independent data partitions is reported in Figure \ref{active} (Accuracy) and Figure \ref{cost} (Cost).

%\vspace{-1em}

From Figures \ref{active} and \ref{cost}, we have the following observations. (i) AMCC significantly outperforms the baseline methods. This global pattern proves the effectiveness of our proposed sample-label-worker triplet selection strategy. (ii) AMCC always outperforms the variant AMCC(rW) on both accuracy and cost, which shows the effectiveness of AMCC in selecting the most suitable workers with low cost and capable of providing reliable annotations for the selected sample-label pairs. (iii) AMCC achieves a better performance than AMCC(rSL) in most cases. This proves that the uncertainty of sample-label pairs helps in reducing the query cost, and AMCC can select useful sample-label pairs. (iv) AMCC(nLC), AMCC(rSL), NAM(A), and SAC ignore label correlations, so they need more queries to achieve the same accuracy as that of AMCC.  (v)  MV(rWS) has always the lowest accuracy, since it randomly selects  sample-label-worker triples; it neither accounts for the uncertainty of samples and label correlations, nor for the individuality of workers in crowdsourcing. (vi) AMCC always takes the lowest cost among all compared methods. Both SAC and NAM(A) select the most reliable workers for annotations, and need a higher budget. (vii) AMCC(nLC), AMCC(rSL), and AMCC all reduce the cost by  selecting  relatively reliable workers with low cost.  From these results, we can conclude that the uncertainty of samples, label correlations, the specialty (including individuality, group commonality, and cost) of workers can jointly reduce the cost of active crowdsourcing learning on multi-label data.

\begin{figure}[h!tbp]
\centering
  \includegraphics[scale=0.32]{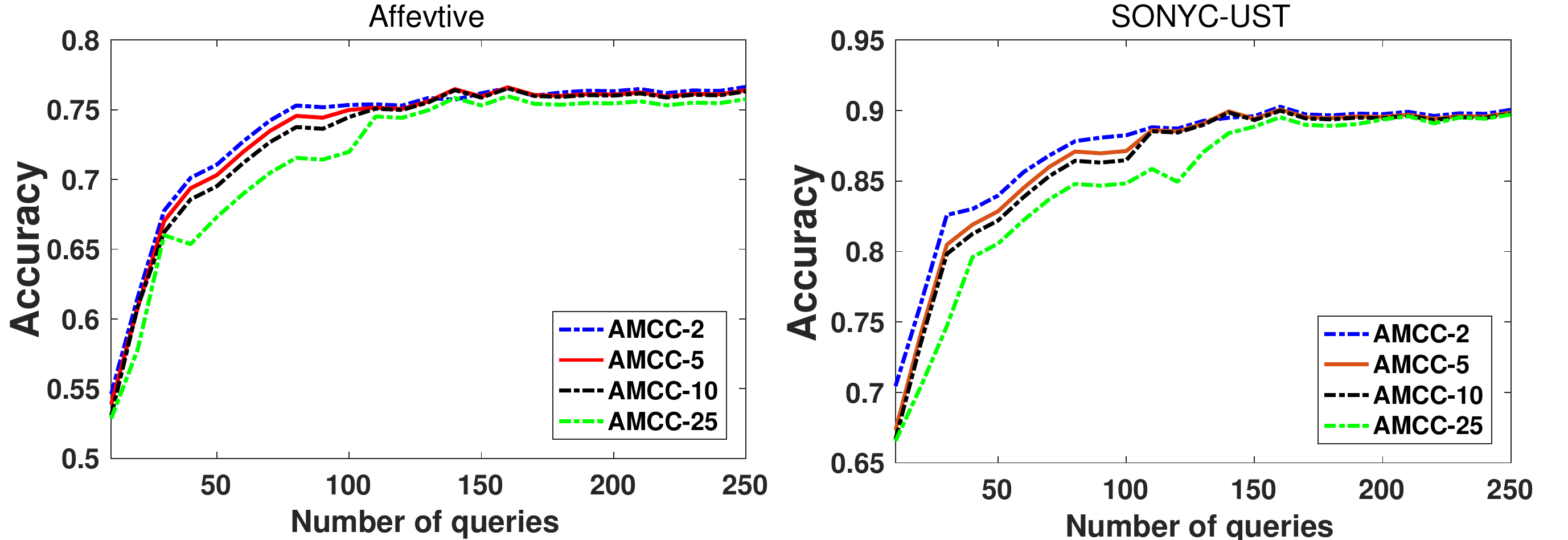}\\
  \caption{The impact of different batch sizes of queries (sample-label pairs) on Affective and SONYC-UST datasets. }\label{batchsize}
\end{figure}

{We further study the impact of query batch size (number of sample-label pairs) on active learning}. We start from 5\% labeled samples of Effective and SONYC-UST datasets until the labeled samples increased to 50\%. We fix the batch size to \{2, 5, 10, 25\} and report the correspond results in Figure \ref{batchsize}. We observe that a smaller batch size generally gives a slightly better performance. That is because a large size provides a batch of samples and labels with more within-redundancy. On the other hand, a smaller batch size asks for more iterations and more computation. For balance, we fix the batch size to 5 for experiments.

\subsection{Robustness with respect to sparse annotations}
In real scenarios of  crowdsourcing, it's common that the majority of workers only annotate few samples, while some workers annotate many.  To evaluate the behavior of  different consensus approaches with respect to annotation sparsity, for each dataset in Table \ref{dataset}, we randomly remove $\{10\%, 20\%, 30\%, 50\%\}$ of the annotations of each worker to generate  sparser annotations, and then adopt the remaining annotations  for the experiments.  In the random removal process, we ensure that each worker annotates at least one sample. The  results are shown in Table \ref{sparseresult}.

\begin{table*}[h!tbp]
\scriptsize
    \caption{Accuracy and  standard deviations of all methods under different ratios of removed annotations. $\circ/\bullet$ indicates AMCC is statistically worse/better than the comparing method, and the significance is assessed using a pairwise $t$-test at 95\% confidence level.}
     \begin{center}
    \begin{tabular}{c|c c c c c c|c}
    \hline
     {Ratios} &MV &C-DS &RA$k$EL-GLAD  &MCMLD  &ML-JMF &NAM   &AMCC\\
    \hline
    & \multicolumn{7}{c}{AppleNakamura}\\
     \cline{2-8}
    \emph{10\%} &$0.929 \pm0.000\bullet$  &$0.931\pm0.012\bullet$   &$0.939\pm0.011\bullet$
    &$0.933\pm0.011\bullet$
    &$0.945\pm0.021\bullet$ &$0.951\pm0.018\bullet$  &$0.958\pm0.017$ \\
    \emph{20\%}   &$0.910\pm0.000\bullet$  &$0.922\pm0.011\bullet$  &$0.924\pm0.011\bullet$
    &$0.924\pm0.021\bullet$
    &$0.934\pm0.011\bullet$  &$0.937\pm0.014\bullet$   &$0.943\pm0.016$ \\
    \emph{30\%}  &$0.834\pm0.000\bullet$  &$0.844\pm0.021\bullet$  &$0.842\pm0.022\bullet$
    &$0.848\pm0.012\bullet$
    &$0.850\pm0.020\bullet$ &$0.854\pm0.024\bullet$  &$0.870\pm0.021$ \\
    \emph{50\%}  &$0.785\pm0.000\bullet$  &$0.794\pm0.020\bullet$  &$0.796\pm0.031\bullet$ &$0.804\pm0.020\bullet$
    &$0.816\pm0.031\bullet$
    &$0.824\pm0.011\bullet$  &$0.841\pm0.009$ \\
    \hline
    & \multicolumn{7}{c}{AppleEkman} \\
     \cline{2-8}
    \emph{10\%} &$0.920\pm0.000\bullet$   &$0.932\pm0.012\bullet$    &$0.936\pm0.015\bullet$
    &$0.926\pm0.017\bullet$
    &$0.942\pm0.015\bullet$
    &$0.956\pm0.011\circ$ &$0.947\pm0.000$ \\
    \emph{20\%}   &$0.902\pm0.000\bullet$     &$0.910\pm0.021\bullet$    &$0.918\pm0.018\bullet$
    &$0.921\pm0.001\bullet$
    &$0.911\pm0.013\bullet$   &$0.923\pm0.017\bullet$    &$0.931\pm0.008$ \\
    \emph{30\%}  &$0.811\pm0.000\bullet$   &$0.852\pm0.026\bullet$
    &$0.859\pm0.006\bullet$
    &$0.863\pm0.016\bullet$  &$0.870\pm0.024\bullet$ &$0.868\pm0.018\bullet$  &$0.881\pm0.002$ \\
    \emph{50\%}  &$0.760\pm0.000\bullet$   &$0.817\pm0.034\bullet$
    &$0.822\pm0.014\bullet$
    &$0.832\pm0.024\bullet$  &$0.838\pm0.027\bullet$ &$0.831\pm0.024\bullet$  &$0.847\pm0.015$ \\
    \hline
    & \multicolumn{7}{c}{LoveNakamura} \\
     \cline{2-8}
    \emph{10\%} &$0.910\pm0.000\bullet$   &$0.922\pm0.005\bullet$
    &$0.927\pm0.015\bullet$
    &$0.932\pm0.005\bullet$   &$0.942\pm0.010\bullet$ &$0.943\pm0.007\bullet$  &$0.951\pm0.001$ \\
    \emph{20\%}   &$0.894\pm0.000\bullet$     &$0.901\pm0.021\bullet$
    &$0.911\pm0.031\bullet$
    &$0.918\pm0.013\bullet$    &$0.922\pm0.020\bullet$   &$0.915\pm0.014\bullet$    &$0.930\pm0.021$ \\
    \emph{30\%}  &$0.817\pm0.000\bullet$   &$0.823\pm0.014\bullet$  &$0.851\pm0.034\bullet$
    &$0.849\pm0.003\bullet$
    &$0.854\pm0.014\bullet$ &$0.834\pm0.016\bullet$  &$0.862\pm0.011$ \\
    \emph{50\%}  &$0.770\pm0.000\bullet$   &$0.794\pm0.011\bullet$  &$0.822\pm0.015\bullet$
    &$0.812\pm0.025\bullet$
    &$0.822\pm0.012\bullet$
    &$0.831\pm0.022\bullet$  &$0.840\pm0.014$ \\
    \hline
    & \multicolumn{7}{c}{LoveEkman} \\
     \cline{2-8}
    \emph{10\%} &$0.910\pm0.000\bullet$   &$0.919\pm0.021\bullet$    &$0.921\pm0.020\bullet$
    &$0.912\pm0.011\bullet$
    &$0.929\pm0.021\bullet$ &$0.939\pm0.014\circ$  &$0.9280\pm0.013$ \\
    \emph{20\%}   &$0.871\pm0.000\bullet$     &$0.881\pm0.011\bullet$
    &$0.872\pm0.021\bullet$
    &$0.895\pm0.011\bullet$  &$0.886\pm0.031\bullet$   &$0.881\pm0.012\bullet$    &$0.892\pm0.011$ \\
    \emph{30\%}  &$0.827\pm0.000\bullet$   &$0.839\pm0.026\bullet$
    &$0.849\pm0.016\bullet$
    &$0.847\pm0.024\bullet$ &$0.852\pm0.031\bullet$ &$0.860\pm0.023\bullet$  &$0.864\pm0.019$ \\
    \emph{50\%}  &$0.734\pm0.000\bullet$   &$0.756\pm0.037\bullet$
    &$0.748\pm0.017\bullet$
    &$0.756\pm0.007\bullet$    &$0.771\pm 0.026\bullet$ &$0.782\pm0.024\bullet$  &$0.803\pm0.020$ \\
    \hline
    & \multicolumn{7}{c}{Movie} \\
     \cline{2-8}
    \emph{10\%} &$0.920\pm0.000\bullet$   &$0.929\pm0.021\bullet$    &$0.931\pm0.011\bullet$
    &$0.932\pm0.021\bullet$
    &$0.929\pm0.028\bullet$ &$0.944\pm0.019\circ$  &$0.938\pm0.014$ \\
    \emph{20\%}   &$0.881\pm0.000\bullet$     &$0.891\pm0.011\bullet$
    &$0.882\pm0.021\bullet$
    &$0.896\pm0.011\bullet$  &$0.896\pm0.031\bullet$   &$0.891\pm0.012\bullet$    &$0.902\pm0.011$ \\
    \emph{30\%}  &$0.837\pm0.000\bullet$   &$0.849\pm0.026\bullet$
    &$0.859\pm0.016\bullet$
    &$0.857\pm0.024\bullet$ &$0.862\pm0.031\bullet$ &$0.866\pm0.023\bullet$  &$0.874\pm0.019$ \\
    \emph{50\%}  &$0.744\pm0.000\bullet$   &$0.766\pm0.037\bullet$
    &$0.758\pm0.017\bullet$
    &$0.766\pm0.007\bullet$    &$0.781\pm 0.026\bullet$ &$0.792\pm0.024\bullet$  &$0.823\pm0.020$ \\
    \hline
    & \multicolumn{7}{c}{Affective} \\
     \cline{2-8}
    \emph{10\%} &$0.701\pm0.000\bullet$   &$0.723\pm0.021\bullet$    &$0.729\pm0.020\bullet$
    &$0.732\pm0.011\bullet$
    &$0.740\pm0.021\bullet$ &$0.744\pm0.014\circ$  &$0.738\pm0.013$ \\
    \emph{20\%}   &$0.671\pm0.000\bullet$     &$0.681\pm0.011\bullet$
    &$0.672\pm0.021\bullet$
    &$0.695\pm0.016\bullet$  &$0.686\pm0.021\bullet$   &$0.681\pm0.022\bullet$    &$0.692\pm0.011$ \\
    \emph{30\%}  &$0.627\pm0.000\bullet$   &$0.639\pm0.016\bullet$
    &$0.649\pm0.016\bullet$
    &$0.647\pm0.026\bullet$ &$0.652\pm0.021\bullet$ &$0.660\pm0.023\bullet$  &$0.664\pm0.019$ \\
    \emph{50\%}  &$0.534\pm0.000\bullet$   &$0.556\pm0.037\bullet$
    &$0.548\pm0.027\bullet$
    &$0.556\pm0.017\bullet$    &$0.571\pm 0.021\bullet$ &$0.582\pm0.024\bullet$  &$0.603\pm0.020$ \\
    \hline
    & \multicolumn{7}{c}{SONYC-UST} \\
     \cline{2-8}
    \emph{10\%} &$0.710\pm0.000\bullet$   &$0.801\pm0.001\bullet$    &$0.811\pm0.010\bullet$
    &$0.800\pm0.010\bullet$
    &$0.855\pm0.011\bullet$ &$0.843\pm0.004\bullet$  &$0.871\pm0.003$ \\
    \emph{20\%}   &$0.681\pm0.000\bullet$     &$0.781\pm0.014\bullet$
    &$0.792\pm0.011\bullet$
    &$0.783\pm0.012\bullet$  &$0.847\pm0.021\bullet$   &$0.838\pm0.012\bullet$    &$0.860\pm0.001$ \\
    \emph{30\%}  &$0.637\pm0.000\bullet$   &$0.723\pm0.015\bullet$
    &$0.749\pm0.015\bullet$
    &$0.752\pm0.016\bullet$ &$0.801\pm0.011\bullet$ &$0.791\pm0.023\bullet$  &$0.831\pm0.012$ \\
    \emph{50\%}  &$0.580\pm0.000\bullet$   &$0.656\pm0.017\bullet$
    &$0.706\pm0.020\bullet$
    &$0.726\pm0.017\bullet$    &$0.731\pm 0.011\bullet$ &$0.741\pm0.024\bullet$  &$0.795\pm0.012$ \\
    \hline
    \end{tabular}
    \end{center}
    \label{sparseresult}
\end{table*}

As the ratio of removed annotations increases, all the methods have a reduced consensus performance. This pattern is expected, since the information collected from workers  gradually diminishes. AMCC almost always outperforms the baselines across all the seven  datasets. MV, C-DS, and RA$k$EL-GLAD  are more sensitive to annotation sparsity. When  30\% of the annotations are removed, MV, C-DS, and RA$k$EL-GLAD show a sharply decrease in accuracy.  This is because MV selects the labels annotated by the majority as the ground truth, and the consensus labels of one instance is less reliable when very few workers annotate it. C-DS and RA$k$EL-GLAD ignore the connections between workers, which leads to an inaccurate consensus. When 50\% of the annotations is removed, AMCC still holds an accuracy $\geq$80\% (except on the Affective dataset) and is more robust to annotation sparsity than the other methods. This is because AMCC considers the commonality of workers at the group level and alleviates the issue of sparse annotations via merging the available annotations in the same group. In addition, it accounts for the individuality of workers. In contrast, the other  methods either ignore the individuality or the commonality.

\subsection{Parameter Sensitivity Analysis}
\begin{figure}[h!tbp]
  \centering
  \includegraphics[scale=0.3]{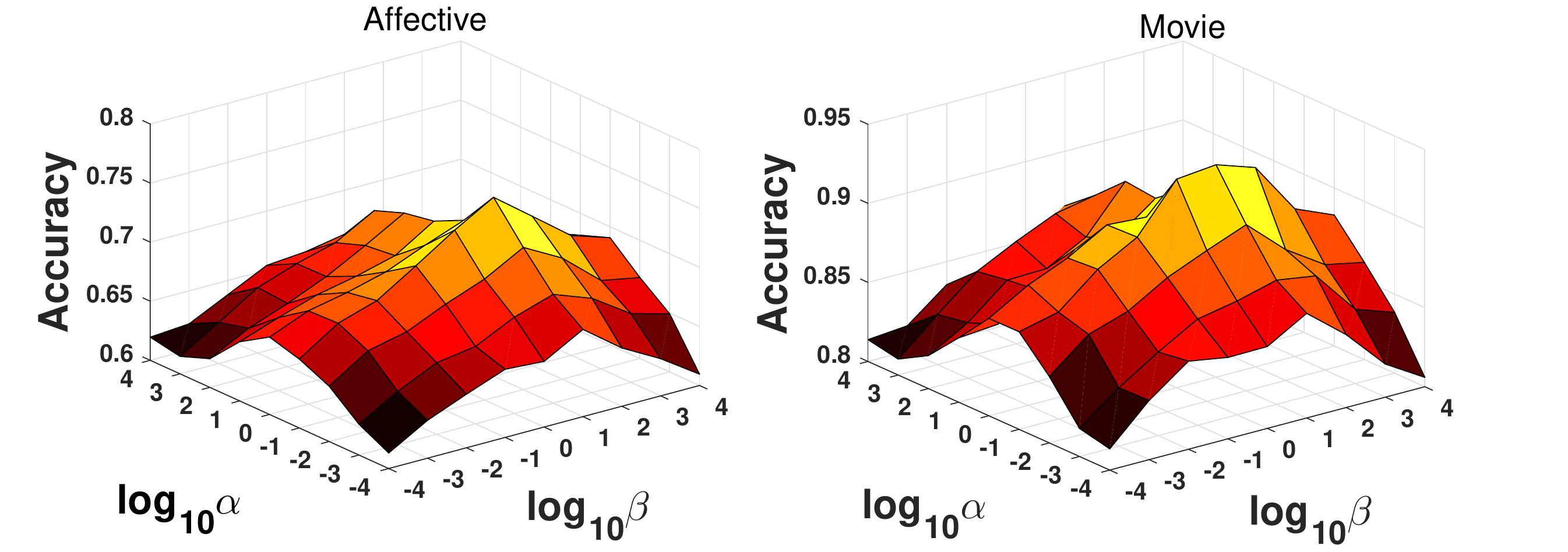}\\
  \caption{Accuracy of AMCC under different combinations of $\alpha$ and $\beta$}\label{alpha-beta}
\end{figure}

\begin{figure}[h!tbp]
  \centering
  \includegraphics[scale=0.3]{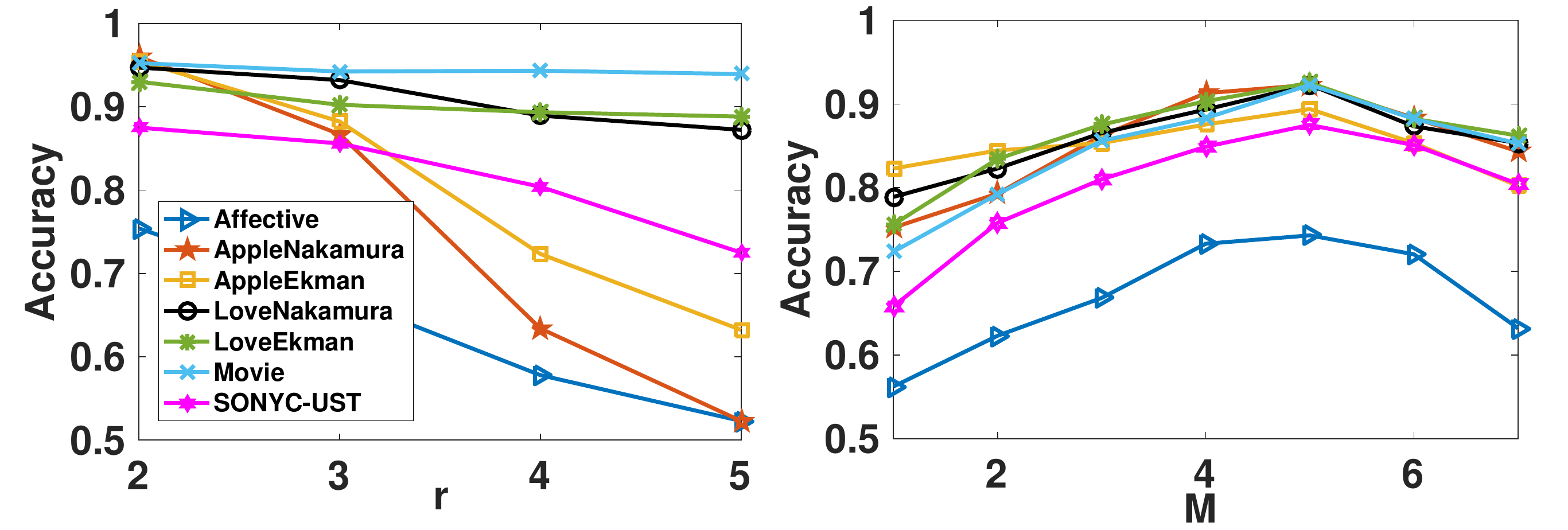}\\
  \caption{Accuracy of AMCC under different values of $r$ and $M$}\label{rm}
\end{figure}

Four input parameters, namely  $\alpha$, $\beta$, $r$, and $M$, may affect the performance of AMCC. We conduct experiments to study the sensitivity of AMCC with respect to these parameters. $\alpha$ and $\beta$ adjust the contribution of label correlations and the commonality between workers in the same group, $r$ scales the weights assigned to groups, and $M$ controls the number of workers' groups. We report in Figure \ref{alpha-beta} the results of AMCC on the Movie and Affective datasets, when $\alpha$ and $\beta$ vary in $\{10^{-4}, 10^{-3}, \cdots, 10^4\}$, and in Figure \ref{rm} those when $r$ vary in $\{2, 3, 4,5\}$ and $M$ in $\{2, 3, \cdots, 7\}$.

When $\alpha$ is fixed, we can see that the accuracy of AMCC  increases first and then decreases, reaching a maximum at $\beta=10$. This is because a small $\beta$ value does not make sufficient use of label correlations, which can often boost the performance of multi-label learning, while a too large $\beta$ overemphasizes label correlations. When $\beta$ is fixed, $\alpha$ values which are too large or too small bring down the accuracy of AMCC. This is because a too small value of $\alpha$ underweighs the commonality of workers, while a too large value of $\alpha$ overweighs the commonality of workers. This pattern indicates that the workers should be properly modeled in crowdsourcing. Based on the above analysis, we set $\alpha=0.1$ and $\beta=10$ in the experiments.

\begin{figure}[h!tbp]
  \includegraphics[scale=0.32]{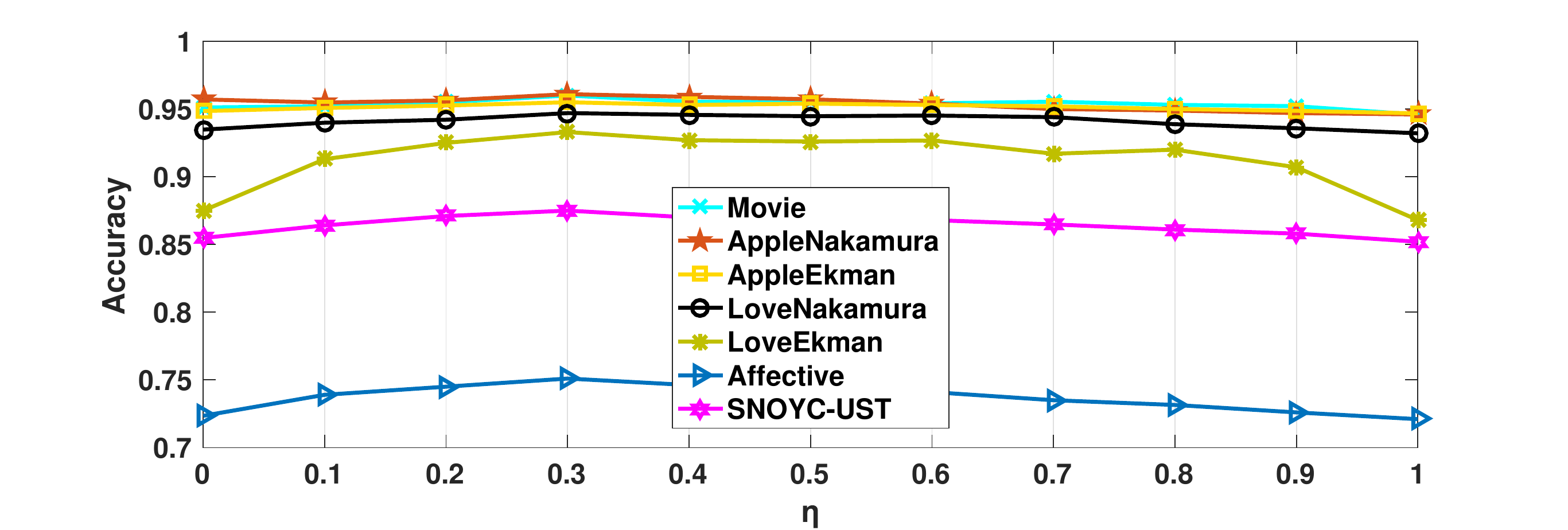}\\
  \caption{Accuracy of AMCC under different values of $\eta$}\label{eta}
\end{figure}
The left sub-figure in Figure \ref{rm} shows that the accuracy of AMCC decreases, or remains steady, as the value of $r$ (power size of weights $\lambda_m$ in Eq. (\ref{eq5})) grows, and achieves the highest value when $r=2$. Therefore, we set $r=2$ in our experiments.

In Figure \ref{rm} (right), we also report the results of AMCC when the number of groups changes. AMCC improves when the number of groups increases. This is because there often exists different types of workers, such as reliable, normal, and sloppy workers, and spammers during crowdsourcing \cite{Tu2018multi}. AMCC boosts its performance  by grouping workers, and by assigning different weights to the groups and workers therein. In practice, based on the study in \cite{hung2013evaluation,kazai2011worker}, there are usually four or five types of workers in the real-world crowdsourcing. Given that, we set $M=5$, which is effective and reasonable for all the experiments.

Figure \ref{eta} shows the Accuracy of AMCC under different values of $\eta$ on the seven datasets in Table \ref{dataset}. We can find that AMCC obtains relatively stable performance when $\eta \in [0.2,0.7]$. When $\eta$ is close to the extreme value (0 or 1), AMCC manifests a reduced performance. That is because AMCC only uses  uncertainty to select sample-label pairs when $\eta=0$, and only uses label correlations to select sample-label pairs when $\eta=1$. In other words, both the uncertainty and label correlations contribute to the sample-label pair selection. Given that, we adopt $\eta=0.3$ for experiments.

\subsection {Convergence analysis}
\begin{figure}[h!tbp]
  \includegraphics[scale=0.32]{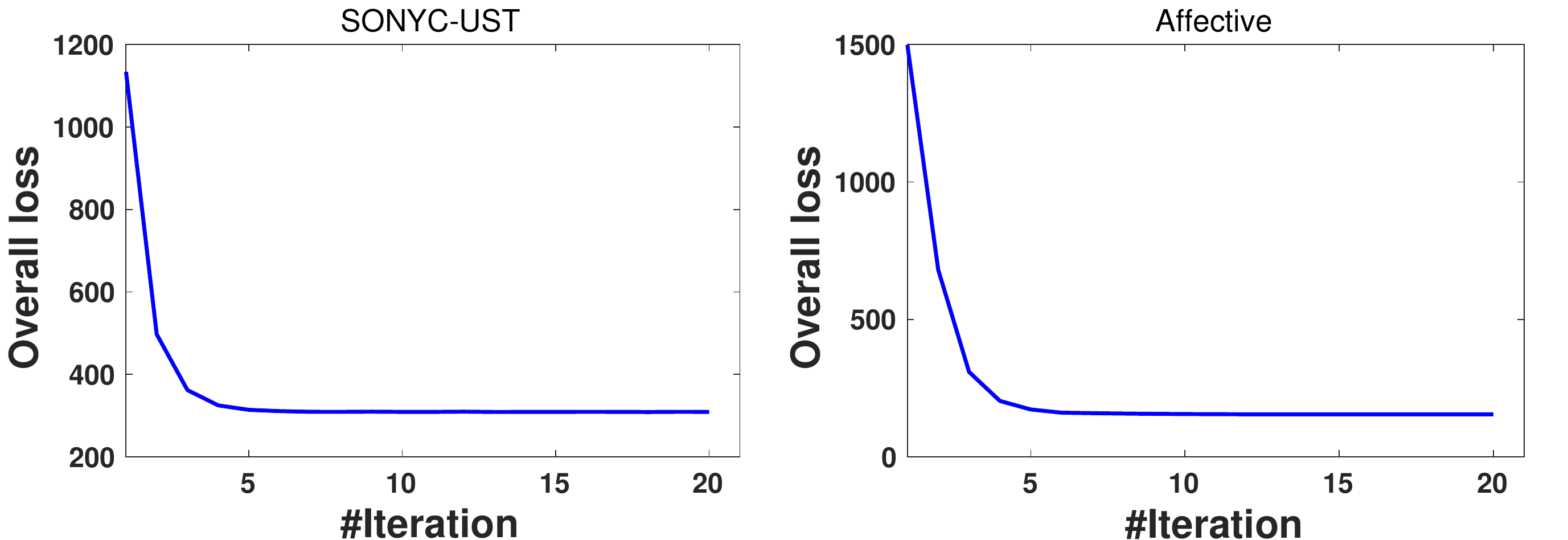}\\
  \caption{Convergence curve of AMCC on the Affective and SONYC-UST datasets. }\label{convergence}
  \vspace{-0.2cm}
\end{figure}

From  the convexity analysis in Section \ref{sec:optimization}, we prove that once $\mu =4L(W-1)\beta$ AMCC will converge. We  plot the loss trend of AMCC in each iteration on the Affective and SONYC-UST datasets in  Figure \ref{convergence}. AMCC quickly converges after five iterations. The overall loss patterns in each iteration on the other datasets give similar patterns. Therefore, AMCC indeed comes to the convergence under the condition $\mu =4L(W-1)\beta$.

\section{Conclusion}
\label{concl}
In this paper, we summarized the challenges of crowd consensus on multi-label data and its conjunction with active learning. For these challenges, we introduced an approach called  Active Multi-label Crowd Consensus (AMCC).  AMCC takes into account the commonality and the individuality of workers, and assumes that workers can be divided into different groups. AMCC reduces the impact of unreliable workers by assigning smaller weights to the groups. To collect reliable annotations with reduced cost, AMCC incorporates a novel active crowdsourcing learning strategy to select sample-label-worker triplets. In the triplet, the selected sample and  label are the most helpful for the consensus model, and the selected worker can reliably annotate the sample with the lowest cost possible. Results on three simulated datasets and {seven} real-world datasets show that AMCC can achieve reliable annotations with low cost, and accurately aggregate  labels for the samples. In addition, AMCC performs well  when the annotations are sparse. The code and datasets will be available at http://mlda.swu.edu.cn/codes.php?name=AMCC.

\section{Acknowledgments}
We appreciate the authors for generous sharing their codes and datasets with us for experiments. This work is supported by Natural Science Foundation of China (61872300 and 61873214), Fundamental Research Funds for the Central Universities (XDJK2019B024), Natural Science Foundation of CQ CSTC (cstc2018jcyjAX0228).

\bibliography{AMCC_bib}
\bibliographystyle{IEEEtran}

\end{document}